\documentclass{article} %
\usepackage{iclr2025_conference,times}

\usepackage{amsmath,amsfonts,bm}

\def\eqref#1{Equation~\ref{#1}}

\def\1{\bm{1}}

\DeclareMathAlphabet{\mathsfit}{\encodingdefault}{\sfdefault}{m}{sl}
\SetMathAlphabet{\mathsfit}{bold}{\encodingdefault}{\sfdefault}{bx}{n}

\usepackage{hyperref}
\usepackage{url}

\usepackage{graphicx}
\usepackage{booktabs}

\usepackage[]{amsmath}                  %
\usepackage[]{mathtools}                %
\usepackage[]{amssymb,amsfonts}
\usepackage{graphicx}
\usepackage{bm}
\usepackage{bbm}

\usepackage[]{xcolor}
\usepackage[]{eucal}

\usepackage[]{xparse}
\usepackage[]{xstring}
\usepackage{etoolbox}
\usepackage[]{xspace}
\usepackage[]{comment}
\usepackage{float}

\usepackage{nicefrac}
\usepackage{amsmath}

\usepackage{multirow}
\usepackage{url}
\usepackage{tcolorbox}

\usepackage{xcolor}

\newcommand{\red}[1]{{\color{black}#1}}

\usepackage{xspace}
\usepackage{multirow}

\newcommand{\tabincell}[2]{
\begin{tabular}{@{}#1@{}}#2\end{tabular}
}

\def\Ours{GenPercept\xspace}

\usepackage[capitalise]{cleveref}

\title{What Matters When Repurposing Diffusion Models for General Dense Perception Tasks?}

\author{\textbf{Guangkai Xu}
~~~~~~
\textbf{Yongtao Ge}
~~~~~~
\textbf{Mingyu Liu}
~~~~~~
\textbf{Chengxiang Fan} \\
\textbf{Kangyang Xie}
~~~~~~
\textbf{Zhiyue Zhao}
~~~~~~
\textbf{Hao Chen}
~~~~~~
\textbf{Chunhua Shen} \\
Zhejiang University
}

\iclrfinalcopy

\begin{document}

\maketitle

\begin{abstract}

Extensive pre-training with large data is indispensable for downstream geometry and semantic visual perception tasks. Thanks to large-scale text-to-image (T2I) pretraining, recent works show promising results by simply fine-tuning T2I diffusion models for 
a few
dense perception tasks. However, several crucial design decisions in this process still lack comprehensive justification, encompassing the necessity of the multi-step diffusion mechanism, training strategy, inference ensemble strategy, and fine-tuning data quality. In this work, we conduct a thorough investigation into critical factors that affect transfer efficiency and performance when using diffusion priors. Our key findings are: 1) High-quality fine-tuning data is paramount for both semantic and geometry perception tasks. 2) 
As a special case
of the diffusion scheduler by setting its hyper-parameters, the multi-step generation can be simplified to a one-step fine-tuning paradigm without any loss of performance.
3) Apart from fine-tuning the diffusion model with only latent space supervision, task-specific 
supervision is beneficial to enhance fine-grained details. These observations culminate in the development of \textbf{GenPercept}, an effective deterministic one-step fine-tuning paradigm tailored for dense visual perception tasks exploiting diffusion priors. Different from the previous multi-step methods, our paradigm %
offers
a much faster inference speed, and can be seamlessly integrated with customized perception decoders and loss functions 
for task-specific 
supervision, which can be critical %
for 
improving the fine-grained details of predictions. Comprehensive experiments on a diverse set of 
dense visual perceptual tasks, including monocular depth estimation, surface normal estimation, image segmentation, and matting, are performed to demonstrate the remarkable adaptability and effectiveness of our proposed method.

\end{abstract}

\section{Introduction}

    Recent studies have explored the transferability of text-to-image (T2I) diffusion models to dense visual perception tasks, such as geometry estimation \citep{ke2023repurposing, lee2023exploiting, geowizard, gui2024depthfm, ye2024stablenormal}, image segmentation \citep{van2024simple, lee2023exploiting}, and inverse rendering \citep{chen2024intrinsicanything, kocsis2024intrinsic, zeng2024rgb}. While these works have demonstrated impressive results by repurposing diffusion models for estimating 
    geometric and semantic dense prediction maps, the critical design choices made in transferring diffusion models to other dense perception tasks still lack comprehensive justification. This makes it challenging to determine the optimal strategy for achieving %
    optimal 
    performance.
    
    For example, \citet{ke2023repurposing} align the visual perception process with the denoising process of Stable Diffusion by fine-tuning all U-Net parameters. They highlight the significance of ``multi-resolution noise'' in the forward diffusion process during training, aiming to obtain clean predictions by gradually removing Gaussian noise. On the other hand, \citet{lee2023exploiting} modify the forward diffusion process by interpolating perception annotations with RGB images instead of using Gaussian noise, and only train the low-rank adaptation (LoRA) \citep{hu2022lora} parameters while keeping the U-Net frozen. To
    our knowledge, the effective components of these approaches have not been thoroughly investigated, and 
    it is unclear which
    design choices 
    contribute most to the success.

    In this work, we examine the design space of repurposing diffusion models for dense visual perception tasks, and %
    attempt to answer 
    the key question: \textbf{What are the important design choices when adapting diffusion models for general dense perception tasks?}

    To answer this question, we rethink the importance of both fine-tuning protocols and fine-tuning data. From the perspective of fine-tuning protocols, we categorize recent methods into two main groups: \textit{stochastic multi-step generation} and \textit{deterministic multi-step generation}. We explore several critical design 
    choices,
    including the diffusion mechanism, key architectural components, training methodologies, and data quality. Our key observations are as follows: \textbf{1)} 
    By %
    setting 
    the hyperparameters of the diffusion scheduler to particular values,
    the multi-step generation can be simplified to a one-step fine-tuning paradigm without any loss of performance.
    \textbf{2)} Strict adherence to traditional diffusion processes %
    appears to be 
    unnecessary. Single-step inference provides similar performance with significantly faster execution. \textbf{3)} High-quality synthetic fine-tuning data is crucial for %
    several
    perception tasks. From the perspective of fine-tuning data quality, we conduct comprehensive dataset ablation studies on both synthetic datasets and real-world datasets.

\begin{figure}[h]
    \centering
    \includegraphics[width=\textwidth]{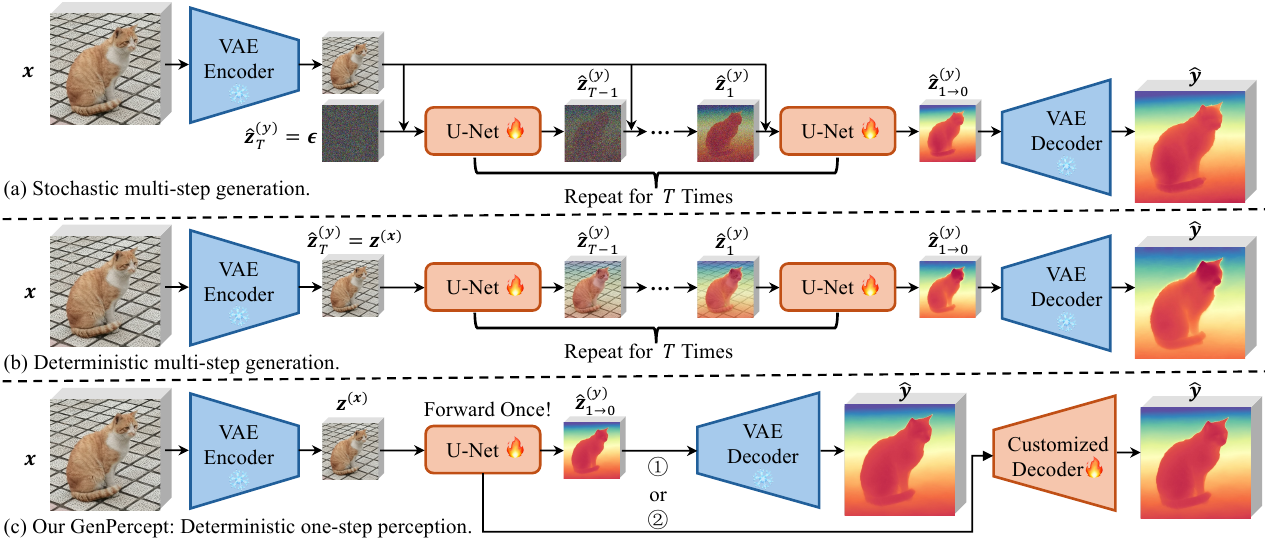}
    \caption{Comparisons of three different pipelines. Our GenPercept enables one-step inference and supports pixel-wise losses and customized decoders to replace the cumbersome VAE decoder. We also extend GenPercept to five dense perception tasks including monocular depth estimation, surface normal estimation, dichotomous image segmentation, semantic segmentation, and image matting. 
    }
    \label{fig: pipeline}
\end{figure}

    Based on the aforementioned observations, we propose \textbf{GenPercept} (see~\cref{fig: pipeline}), a deterministic fine-tuning paradigm featuring a remarkably simple one-step inference pipeline, an optional customized decoder, and an easily adaptable %
    pixel-wise 
    loss. We conduct extensive quantitative and qualitative experiments on a wide range of %
    visual dense perception tasks, including monocular depth estimation, surface normal estimation, image segmentation, and matting to demonstrate the effectiveness and 
    generalization capability 
    of our method.
    
    In conclusion, our contributions can be summarized as follows: \textbf{1)} We systematically analyze the design space of %
    fine-tuning protocols, considering both model architecture and dataset selection, through comprehensive ablation studies. \textbf{2)} Based on these insights, we propose \Ours, a simple paradigm that harnesses the power of the pre-trained UNet from diffusion models for generalizable dense visual perception tasks.

\section{preliminary}
\label{sec: preliminary}

\newcommand{\image}{\mathbf{x}}
\newcommand{\gt}{\mathbf{y}}
\newcommand{\latent}{\mathbf{z}}
\newcommand{\latentgt}{\latent^{(\gt)}}
\newcommand{\latenthatgt}{\hat{\latent}^{(\gt)}}
\newcommand{\latentimage}{\latent^{(\image)}}
\newcommand{\latenthatimage}{\hat{\latent}^{(\image)}}
\newcommand{\noise}{\bm{\epsilon}}
\newcommand{\denoiser}{\bm{v}_{\theta}}
\newcommand{\prednoise}{\hat{\bm{\noise}}}
\newcommand{\denoiserlong}{\denoiser(\latentgt_t, \latentimage, t)}
\newcommand{\encoder}{\mathcal{E}}
\newcommand{\decoder}{\mathcal{D}}

\red{
We take the latent diffusion model as an example. To model the data distribution, the idea of the diffusion model \citep{Rombach_2022_CVPR, Chen2023PixArtFT, song2020denoising, ho2020denoising} is to randomly sample a noise $\latentgt_{T}\sim \mathcal{N}(\mathbf{0}, \mathbf{I})$ and sequentially denoise it into a $\latentgt_0$, which is distributed according to the data. In the forward diffusion process, $\latentgt_t$ is sampled by $\latentgt_t = \sqrt{\bar{\alpha}_t} \latentgt + \sqrt{1 - \bar{\alpha}_t} \noise$, where $\noise\sim \mathcal{N}(\mathbf{0}, \mathbf{I})$, and $\bar{\alpha}_t = \prod_{s=1}^{t}{(1 - \beta_s)}$. The variance schedule $\{ \beta_t \in (0, 1)\}_{t=1}^T$ is interpolated between $\beta_{start}$ and $\beta_{end}$ with $T$ steps, where larger values of $(\beta_{start}, \beta_{end})$ correspond to smaller $\bar{\alpha}_t$ values, \textit{i.e.}, smaller proportion of noise. In the reverse process, \citet{salimans2021progressive} use a ``v-prediction'' objective, where a denoiser $\denoiser$ minimizes the following:
\begin{equation}
\label{eq: noise_blending_loss}
\begin{gathered}
    \mathcal{L} = \mathbb{E}_{\latentgt, \noise \sim \mathcal{N}(0,I),t \sim \mathcal{U}(T)} \left\| (\sqrt{\bar{\alpha}_t} \noise - \sqrt{1 - \bar{\alpha}_t} \latentgt ) - \denoiser(\latent_t, t) \right\|^2_2.
\end{gathered}
\end{equation}

To fully leverage the pre-trained prior of the diffusion models for dense prediction tasks, previous works have reformulated these tasks as a multi-step denoising process, especially on monocular depth estimation. Given a data pair ($\latentimage$, $\latentgt$) where $\latentimage$ is the observation and $\latentgt$ is the prediction target, \textit{stochastic multi-step generation} methods \citep{ke2023repurposing, geowizard, gui2024depthfm} such as Marigold \citep{ke2023repurposing} add $\latentimage$ as an additional input to the denoiser $\denoiser$, and use $\denoiser(\latentgt_t, \latentimage, t)$ to predict $\latentgt$. By contrast, \textit{deterministic multi-step generation} methods such as DMP  \citep{lee2023exploiting} take the observation $\latentimage$ as a deterministic noise and compose $\latentgt_t$ as a blend between $\latentimage$ and $\latentgt$:

\begin{equation}
\label{eq: rgb_noise_blending}
\begin{gathered}
    \latent_t = \latentgt_t = \sqrt{\bar{\alpha}_t} \latentgt + \sqrt{1 - \bar{\alpha}_t} \latentimage, ~~t = [1, ..., T], \\
\end{gathered}
\end{equation}

The denoising process and forward diffusion process of these two categories are illustrated in \cref{fig: pipeline} and \cref{fig: viz_beta_change_normal} (a). We offer detailed formulations in the supplementary material.
}

\begin{figure}[t]
    \centering
    \includegraphics[width=\textwidth]{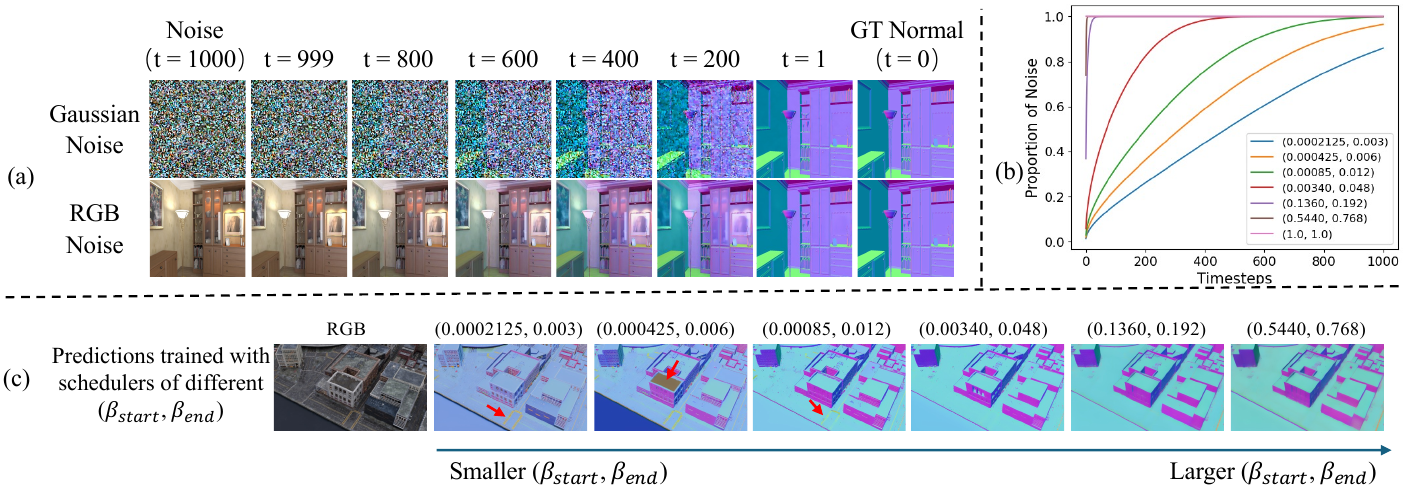}
    \caption{Illustration of different noise forms and proportions in the forward diffusion process. (a) Visualization of interpolating ground-truth labels with Gaussian noise and RGB noise. (b) The relationship between the noise proportion $\sqrt{\bar{\alpha}_t}$ and the ($\beta_{start}$, $\beta_{end}$) hyperparameters. (c) 
    Small 
    $(\beta_{start}, \beta_{end})$ values during the training of deterministic multi-step generation 
    tend to 
    lead to unclean estimation, which %
    contains some 
    RGB information. %
    Enlarging 
    them may alleviate this issue.}
    \label{fig: viz_beta_change_normal}
\end{figure}

\section{Diffusion Models for Visual Perception Tasks}

 In this section, we explore the necessity and highlight the findings of the multi-step diffusion mechanism, 
 the architectural components, training strategy, and fine-tuning data quality. We select the stochastic method Marigold \citep{ke2023repurposing} and the deterministic method DMP \citep{lee2023exploiting} as our baseline methods. The default experimental setting %
 here 
 is similar to \citet{ke2023repurposing} and can be found in the supplementary material.

\subsection{The Form and Proportion of Noise in the Forward Diffusion Process}
\label{sec: form_proportion_of_noise}

\red{

For the training process of Marigold and DMP, the timestep $t$ is sampled to control the proportion of noise added to the ground truth, and the network is trained to recover a clean ground truth from a noisy latent. For smaller timesteps %
such as 
``$t = 200$'', as illustrated in \cref{fig: viz_beta_change_normal}(a), the input to the network retains significant ground truth information (\textit{e.g.}, the purple color of the normal). 
Therefore, we hypothesize that
\textit{a certain level of ground-truth label information being part of the input} 
makes %
the network 
comparatively be easier to recover the clean ground truth latent than %
the case of 
the absence of any ground truth information during training.
This can 
limit the network's capacity to learn comprehensive knowledge and lead to %
unsatisfactory 
performance, as it is  known that 
networks can become lazy that tend to exploit  ``shortcuts''---the ground truth labels in the input in our case. On the other hand, for inference there is no such ground truth available, causing disparity of input signals  between inference and training.
We note that 
T2I tasks may suffer less due to their stochastic nature, namely, converting a text prompt to a generated image is one-to-many mapping process.

To alleviate this issue, we attempt to control the blending proportion by changing the ($\beta_{start}$, $\beta_{end}$) values of the diffusion model's DDPM scheduler.
As shown in \cref{fig: viz_beta_change_normal}(b) and Fig. 2 of the supplementary, training with a ($\beta_{start}$, $\beta_{end}$) value of (1.0, 1.0) achieves the best rank performance for both Gaussian noise and RGB noise,
which is %
demonstrated  
in \cref{tab: form_and_proportion_of_noise}.
Rather than achieving consistent performance improvement while increasing the noise proportion, we observed that Marigold's performance begins to be slightly unstable when the ($\beta_{start}$, $\beta_{end}$) values are sufficiently high. Experiments of varying the random seed during both the training and inference process are conducted to rule out the influence of randomness.
The results show that we may not be able to exactly find a set of unique values for ($\beta_{start}$, $\beta_{end}$) to achieve the best accuracy, as the final accuracy can be  
affected by %
a few other 
factors besides the aforementioned one.

Additionally, when ($\beta_{start}$, $\beta_{end}$) are equal to 1, the noise proportion $\sqrt{\bar{\alpha}_t}$ is equal to 0, and the formulation of DMP can be derived from \cref{eq: noise_blending_loss} and \cref{eq: rgb_noise_blending} as follows.
\begin{equation}
\label{eq: noise_blending_loss_beta_1}
\begin{gathered}
    \latent_t = \latentgt_t = \sqrt{\bar{\alpha}_t} \latentgt + \sqrt{1 - \bar{\alpha}_t} \latentimage = \latentimage, ~~t = [1, ..., T], \\
    \mathcal{L}
    = \mathbb{E}_{\latentgt, \noise \sim \mathcal{N}(0,I),t \sim \mathcal{U}(T)} \left\| - \latentgt  - \denoiser(\latent_t, t) \right\|^2_2.
\end{gathered}
\end{equation}

In this case, the output of the denoiser $\denoiser(\cdot, \cdot)$ is enforced to learn the negative value of the ground truth latent for each step, and the multi-step denoising is equivalent to the single-step denoising.  We propose to reduce the DDIM steps of DMP to one and call it ``deterministic one-step perception''. The resulting inference 
can be significantly faster, 
with performance remaining 
almost unchanged.
We %
name this 
as ``our baseline'' for the subsequent analysis.
}

\begin{table*}[t]
  \centering
  \caption{Comprehensive quantitative comparisons about the impact of noise forms and proportions in the forward diffusion process on monocular depth estimation. Visualizations of different noise forms and the effect of ($\beta_{start}$, $\beta_{end}$) values are shown in \cref{fig: viz_beta_change_normal}. The performance of DMP improves steadily, while Marigold shows initial improvements followed by a decline. When $\beta_{start}$ and $\beta_{end}$ are equal to 1, the inference process can be reduced to one step without compromising performance. ``Rank'' means the average rank of ten evaluation performance (smaller is better).
  }
\resizebox{\linewidth}{!}{%
  \begin{tabular}{@{}r|c|c|c|c|lr|lr|lr|lr|lr| l@{}}
    \toprule
	
	\multirow{2}{*}{Type} & Noise & Multi-res & \multirow{2}{*}{Steps} & \multirow{2}{*}{($\beta_{start}$, $\beta_{end}$)} & \multicolumn{2}{c|}{KITTI}  & \multicolumn{2}{c|}{NYU} & \multicolumn{2}{c|}{ScanNet} 
 & \multicolumn{2}{c|}{DIODE} & \multicolumn{2}{c|}{ETH3D} & \multirow{2}{*}{Rank$\downarrow$}\\
	
    \cline{6-15}
	
    & Form & Noise &  &  &  AbsRel$\downarrow$ & $\delta_1$$\uparrow$ & AbsRel$\downarrow$ & $\delta_1$$\uparrow$ & 
    AbsRel$\downarrow$ & $\delta_1$$\uparrow$ & AbsRel$\downarrow$ & $\delta_1$$\uparrow$ & AbsRel$\downarrow$ & $\delta_1$$\uparrow$ \\

    \hline
    
    Marigold    & Gaussian  & $\checkmark$  & 10  & (0.0002125, 0.003)
                & 0.358  & 0.462
     		& 0.297  & 0.555
     		& 0.246  & 0.625
                & 0.494  & 0.565
                & 0.267  & 0.640
                & 7.0
     		\\
    Marigold    & Gaussian  & $\checkmark$  & 10  & (0.000425, 0.006)
                & 0.122  & 0.854
     		& 0.106  & 0.887
     		& 0.136  & 0.829
                & 0.345  & 0.716
                & 0.086  & 0.927
                & 5.8
     		\\
    baseline Marigold    & Gaussian  & $\checkmark$  & 10  & (0.00085, 0.012)
                & 0.099  & 0.909
     		& 0.063  & 0.956  
     		& 0.075  & 0.937  
                & 0.316  & 0.764  
                & 0.075  & 0.947  
                & 3.6
     		\\
    Marigold    & Gaussian  & $\checkmark$  & 10  & (0.0034, 0.048)
                & 0.100  & 0.906 
     		& 0.057  & 0.963  
     		& 0.063  & 0.957  
                & 0.308  & 0.768  
                & 0.074  & 0.948  
                & 2.3
     		\\
    Marigold    & Gaussian  & $\checkmark$  & 10  & (0.1360, 0.192)
                & 0.119  & 0.861  
     		& 0.058  & 0.963  
     		& 0.061  & 0.961  
                & 0.315  & 0.760  
                & 0.073  & 0.950  
                & 2.8
     		\\
    Marigold    & Gaussian  & $\checkmark$  & 10  & (0.5440, 0.768)
                & 0.124  & 0.852  
     		& 0.060  & 0.961  
     		& 0.064  & 0.958  
                & 0.322  & 0.749  
                & 0.079  & 0.943  
                & 4.7
     		\\
    Marigold 	& Gaussian  & $\checkmark$  & 10  & (1.0, 1.0)
                & 0.104  & 0.897
     		& 0.055  & 0.965
     		& 0.059  & 0.962
                & 0.312  & 0.762
                & 0.069  & 0.955
                & 1.7
     		\\
    \hline
    
    Marigold 	& Gaussian  & $\checkmark$  & 1  & (1.0, 1.0)
                & 0.104  & 0.897
     		& 0.055  & 0.965
     		& 0.059  & 0.962
                & 0.312  & 0.762
                & 0.069  & 0.955
                & -
     		\\

    \hline
    
    Marigold    & Gaussian  & $\times$  & 10  & (0.0002125, 0.003)
                & 0.587  & 0.255  
     		& 0.337  & 0.490  
     		& 0.257  & 0.604  
                & 0.600  & 0.469  
                & 0.372  & 0.503  
                & 7.0
     		\\
    Marigold    & Gaussian  & $\times$  & 10  & (0.000425, 0.006)
                & 0.536  & 0.289  
     		& 0.313  & 0.527  
     		& 0.248  & 0.621  
                & 0.565  & 0.499  
                & 0.328  & 0.575  
                & 6.0
     		\\
    baseline Marigold    & Gaussian  & $\times$  & 10  & (0.00085, 0.012)
                & 0.153  & 0.807  
     		& 0.162  & 0.802
     		& 0.187  & 0.737
                & 0.411  & 0.641  
                & 0.157  & 0.826  
                & 5.0
     		\\
    Marigold    & Gaussian  & $\times$  & 10  & (0.0034, 0.048)
                & 0.101  & 0.907  
     		& 0.058  & 0.963  
     		& 0.066  & 0.954  
                & 0.309  & 0.765  
                & 0.074  & 0.950  
                & 2.4
     		\\
    Marigold    & Gaussian  & $\times$  & 10  & (0.1360, 0.192)
                & 0.115  & 0.870  
     		& 0.056  & 0.965  
     		& 0.060  & 0.961 
                & 0.313  & 0.763  
                & 0.072  & 0.953  
                & 2.3
     		\\
    Marigold    & Gaussian  & $\times$  & 10  & (0.5440, 0.768)
                & 0.124  & 0.848  
     		& 0.059  & 0.963  
     		& 0.063  & 0.958  
                & 0.318  & 0.752  
                & 0.077  & 0.946  
                & 3.7
     		\\
    Marigold 	& Gaussian  & $\times$  & 10  & (1.0, 1.0)
                & 0.102  & 0.901
     		& 0.054  & 0.966
     		& 0.059  & 0.962
                & 0.312  & 0.762
                & 0.071  & 0.955
                & 1.5
     		\\
    \hline
    
    Marigold 	& Gaussian  & $\times$  & 1  & (1.0, 1.0)
                & 0.102  & 0.901
     		& 0.054  & 0.966
     		& 0.059  & 0.962
                & 0.312  & 0.762
                & 0.071  & 0.955
                & -
     		\\
    
    \hline
    DMP 	    & RGB & $\times$  & 10  & (0.0002125, 0.003)
                & 0.476  & 0.336  
     		& 0.267  & 0.601 
     		& 0.216  & 0.677  
                & 0.457  & 0.588  
                & 0.185  & 0.757  
                & 6.9
     		\\
    DMP 	    & RGB & $\times$  & 10  & (0.000425, 0.006)
                & 0.265  & 0.630  
     		& 0.201  & 0.072  
     		& 0.195  & 0.717  
                & 0.386  & 0.674  
                & 0.116  & 0.880  
                & 6.1
     		\\
    baseline DMP 	    & RGB & $\times$  & 10  & (0.00085, 0.012)
                & 0.134  & 0.837
     		& 0.117  & 0.871  
     		& 0.147  & 0.808  
                & 0.353  & 0.721  
                & 0.093  & 0.919  
                & 5.0
     		\\
    DMP 	    & RGB & $\times$  & 10  & (0.0034, 0.048)
                & 0.107  & 0.890  
     		& 0.077  & 0.939  
     		& 0.087  & 0.923  
                & 0.318  & 0.766  
                & 0.078  & 0.940  
                & 3.8
     		\\
    DMP 	    & RGB & $\times$  & 10  & (0.1360, 0.192)
                & 0.107  & 0.890  
     		& 0.063  & 0.959  
     		& 0.068  & 0.955  
                & 0.305  & 0.773  
                & 0.073  & 0.948  
                & 2.2
     		\\
    DMP 	    & RGB & $\times$  & 10  & (0.5440, 0.768)
                & 0.106  & 0.897  
     		& 0.061  & 0.959  
     		& 0.066  & 0.952  
                & 0.309  & 0.768  
                & 0.075  & 0.945  
                & 2.3
     		\\

    DMP 	    & RGB & $\times$  & 10  & (1.0, 1.0)
                & 0.100  & 0.902  
     		& 0.053  & 0.966  
     		& 0.059  & 0.961  
                & 0.309  & 0.768  
                & 0.068  & 0.956  
                & 1.2
     		\\
    
    \hline

    Our baseline   & RGB  & $\times$  & 1  & (1.0, 1.0) 
                & 0.100  & 0.902  
     		& 0.053  & 0.966  
     		& 0.059  & 0.961  
                & 0.309  & 0.768  
                & 0.068  & 0.956  
                & -
     		\\
     
    \bottomrule
  \end{tabular}
  }
  \label{tab: form_and_proportion_of_noise}
\end{table*}

\begin{tcolorbox}[
colframe=black,
arc=4pt,
boxsep=1pt,
]
    \paragraph{\textbf{\textit{Finding} 1.}} 
    \red{By 
    setting 
    the ($\beta_{start}$, ${\beta_{end}}$) values to 1, the multi-step generation 
    is 
    simplified to a one-step fine-tuning paradigm without any loss of performance in both stochastic and deterministic methods, \textit{e.g.}, Marigold \citep{ke2023repurposing} and DMP \citep{lee2023exploiting} respectively.}
\end{tcolorbox}

\subsection{Where Does the Rich Visual Knowledge Reside in Diffusion Models?}

Based on the baseline we proposed in \S \ref{sec: form_proportion_of_noise}, we conduct detailed ablation studies to thoroughly investigate the necessity of each component of Stable Diffusion. Results are reported in \cref{tab: ablation_components}.

\begin{table*}[h]
  \centering
  \caption{Explorations on the impact of the Stable Diffusion components on depth estimation. Customized decoders and losses can also enable inference acceleration and performance improvement.}
\resizebox{\linewidth}{!}{%
  \begin{tabular}{@{}r|c|lr|lr|lr|lr|lr@{}}
    \toprule
	
	\multirow{2}{*}{Setting} & \multirow{2}{*}{Loss} & \multicolumn{2}{c|}{KITTI}  & \multicolumn{2}{c|}{NYU} & \multicolumn{2}{c|}{ScanNet} 
 & \multicolumn{2}{c|}{DIODE} & \multicolumn{2}{c}{ETH3D}\\
	
    \cline{3-12}
	
    & & AbsRel$\downarrow$ & $\delta_1$$\uparrow$ & AbsRel$\downarrow$ & $\delta_1$$\uparrow$ & 
    AbsRel$\downarrow$ & $\delta_1$$\uparrow$ & AbsRel$\downarrow$ & $\delta_1$$\uparrow$ & AbsRel$\downarrow$ & $\delta_1$$\uparrow$ \\

    \hline

    Our baseline & MSE (Latent)
                & 0.100  & 0.902  
     		& 0.053  & 0.966  
     		& 0.059  & 0.961  
                & 0.309  & 0.768  
                & 0.068  & 0.956  
     		\\
    \hline
    Train U-Net from scratch   	&  MSE (Latent)
                & 0.219  & 0.650  
     		& 0.186  & 0.736  
     		& 0.183  & 0.729  
                & 0.426  & 0.614  
                & 0.185  & 0.741  
     		\\
    \hline
    Train VAE decoder from scratch  & MSE (Image)
                & 0.096  & 0.916
     		& 0.055  & 0.964
     		& 0.058  & 0.964
                & 0.302  & 0.759
                & 0.071  & 0.950  
     		\\
    \hline
    Baseline + Image MSE loss & MSE (Image)
                & 0.097  & 0.915   
     		& 0.054  & 0.964  
     		& 0.059  & 0.964  
                & 0.305  & 0.760  
                & 0.071  & 0.953  
     		\\
    \hline
    \multirow{2}{*}{Baseline + Image customized loss} & MSE + SSI + 
                & \multirow{2}{*}{0.094}  & \multirow{2}{*}{0.923}  
     		& \multirow{2}{*}{0.052}  & \multirow{2}{*}{0.966}  
     		& \multirow{2}{*}{0.056}  & \multirow{2}{*}{0.965}  
                & \multirow{2}{*}{0.302}  & \multirow{2}{*}{0.767}  
                & \multirow{2}{*}{0.066}  & \multirow{2}{*}{0.967}  
     		\\

      & Grad. (Image)
                &   &   
     		&   &   
     		&   &   
                &   &   
                &   &   
     		\\
    \hline
    Train DPT decoder from scratch  & MSE (Image)
                & 0.099  & 0.912
     		& 0.055  & 0.964
     		& 0.058  & 0.963  
                & 0.302  & 0.759  
                & 0.069  & 0.956  
     		\\
    \bottomrule
  \end{tabular}
  }
  \label{tab: ablation_components}
\end{table*}

\noindent\textbf{Denoiser.}
We reinitialize the U-Net parameters and train the network from scratch on the same datasets. Without prior knowledge of large data from LAION-5B, the network performs poorly and loses the generalization capability. This indicates that most of the prior knowledge is stored in the U-Net %
module. 

\noindent\textbf{VAE AutoEncoder.}
The VAE encoder's original architecture is kept intact to maintain the consistency of the encoding process. For the VAE decoder, we train it from scratch with image pixel MSE loss. Without pre-trained parameters of the VAE decoder, it still performs well.

\noindent\textbf{Customized Head and Loss.}
The deterministic one-step perception pipeline enables customized heads and loss functions. By utilizing a DPT decoder \citep{ranftl2021_dpt} and the loss functions of DepthAnythingv2 \citep{yang2024depthv2}, we can implement a lightweight decoder that supervises pixel-wise information at a higher resolution rather than latent features at 1$/$8 resolution. This approach can accelerate inference times and enhance the acquisition of fine-grained details.

\begin{tcolorbox}[
colframe=black,
arc=4pt,
boxsep=1pt,
]
    \paragraph{\textbf{\textit{Finding} 2.}} 
    The primary perceptual prior knowledge of diffusion models is encapsulated within the
    U-Net of the diffusion model.
    Customized heads and loss functions %
    offers flexibility and may lead to faster 
    inference speed and 
    improved results. 
\end{tcolorbox}

\subsection{What About the Timesteps and Text Prompts?}

The timesteps and text prompts are crucial elements in utilizing the Stable Diffusion model to generate diverse images. We conducted ablation studies to investigate their significance. The results reported in \cref{tab: ablation_timesteps_text_inputs} indicate a negligible difference between various settings. Owing to the inherent certainty associated with visual perception tasks, the diversity typically offered by the textual inputs appears to be unnecessary. Similarly, the utility of timesteps is reduced, as the single-step paradigm does not require progressive denoising.

\begin{table*}[h]
  \centering
  \caption{Quantitative comparisons among different timesteps and text prompts on depth estimation.
  }
\resizebox{\linewidth}{!}{%
  \begin{tabular}{@{}r|c|c|lr|lr|lr|lr|lr@{}}
    \toprule
	
	\multirow{2}{*}{Setting} & \multirow{2}{*}{Text Prompt} & Train / Infer & \multicolumn{2}{c|}{KITTI}  & \multicolumn{2}{c|}{NYU} & \multicolumn{2}{c|}{ScanNet} 
 & \multicolumn{2}{c|}{DIODE} & \multicolumn{2}{c}{ETH3D}\\
	
    \cline{4-13}
	
    &  & Timesteps &  AbsRel$\downarrow$ & $\delta_1$$\uparrow$ & AbsRel$\downarrow$ & $\delta_1$$\uparrow$ & 
    AbsRel$\downarrow$ & $\delta_1$$\uparrow$ & AbsRel$\downarrow$ & $\delta_1$$\uparrow$ & AbsRel$\downarrow$ & $\delta_1$$\uparrow$ \\

    \hline

    Our baseline & ``'' & Random / 1
                & 0.100  & 0.902  
     		& 0.053  & 0.966  
     		& 0.059  & 0.961  
                & 0.309  & 0.768  
                & 0.068  & 0.956  
     		\\

    \hline

    Valid text input & ``A high quality RGB image'' & Random / 1
                & 0.101  & 0.900
     		& 0.053  & 0.967
     		& 0.058  & 0.964
                & 0.312  & 0.762
                & 0.070  & 0.954
     		\\

    Random text input & ``\textnormal{F3@qV!k2*\#Zp\^{}n\%1Lz}'' & Random / 1
                & 0.099  & 0.904
     		& 0.054  & 0.965
     		& 0.059  & 0.963
                & 0.311  & 0.763
                & 0.069  & 0.955
     		\\
    \hline

    Timestep1 & ``'' & 1 / 1
                & 0.100  & 0.906
     		& 0.054  & 0.965
     		& 0.060  & 0.961
                & 0.304  & 0.769
                & 0.069  & 0.956
     		\\

    Timestep500 & ``'' & 500 / 500
                & 0.102  & 0.897
     		& 0.053  & 0.966
     		& 0.059  & 0.961
                & 0.307  & 0.765
                & 0.068  & 0.956
     		\\

    Timestep900 & ``'' & 900 / 900
                & 0.105  & 0.891
     		& 0.054  & 0.966
     		& 0.058  & 0.964
                & 0.309  & 0.762
                & 0.068  & 0.953
     		\\
    
    \bottomrule
  \end{tabular}
  }
  \label{tab: ablation_timesteps_text_inputs}
\end{table*}

\begin{tcolorbox}[
colframe=black,
arc=4pt,
boxsep=1pt,
]
    \paragraph{\textbf{\textit{Finding} 3.}} The timesteps and text prompts of diffusion models are negligible for the performance of visual perception tasks.
\end{tcolorbox}

\subsection{How to Leverage the U-Net's Prior Knowledge?}

The significance of the denoiser cannot be overstated. However, the strategies for its utilization are worth a careful 
study.
Should we freeze the denoiser, utilize its intermediate features, and merely fine-tune the decoder for specific tasks? Alternatively, 
can 
we employ LoRA \citep{hu2022lora} instead of extensively fine-tuning the entire denoiser? Unfortunately, the evidence suggests that neither approach is ideal. As illustrated in \cref{tab: ablation_freeze_finetune_lora}, freezing the denoiser significantly compromises performance. Although incorporating LoRA offers some advantages, it %
may 
not fully leverage the potential of denoiser, especially with regular LoRA ranks of 4 and 16. This limitation likely stems from the substantial differences between the noise-to-image denoising process and the image-to-perception prediction task.

\begin{table*}[t]
  \centering
  \caption{Explorations on the paradigms to leverage U-Net's prior knowledge on depth estimation.}
\resizebox{\linewidth}{!}{%
  \begin{tabular}{@{}r|c|lr|lr|lr|lr|lr@{}}
    \toprule
	
	\multirow{2}{*}{Setting} & LoRA & \multicolumn{2}{c|}{KITTI}  & \multicolumn{2}{c|}{NYU} & \multicolumn{2}{c|}{ScanNet} 
 & \multicolumn{2}{c|}{DIODE} & \multicolumn{2}{c}{ETH3D}\\
	
    \cline{3-12}
	
    & Rank & AbsRel$\downarrow$ & $\delta_1$$\uparrow$ & AbsRel$\downarrow$ & $\delta_1$$\uparrow$ & 
    AbsRel$\downarrow$ & $\delta_1$$\uparrow$ & AbsRel$\downarrow$ & $\delta_1$$\uparrow$ & AbsRel$\downarrow$ & $\delta_1$$\uparrow$ \\

    \hline

    Our baseline & -
                & 0.100  & 0.902  
     		& 0.053  & 0.966  
     		& 0.059  & 0.961  
                & 0.309  & 0.768  
                & 0.068  & 0.956  
     		\\
       
    Freeze U-Net + Train DPT decoder & -
                & 0.144  & 0.803
     		& 0.086  & 0.931   
     		& 0.097  & 0.911
                & 0.309  & 0.768  
                & 0.068  & 0.956
     		\\

    Train U-Net with LoRA & 4
                & 0.211  & 0.644  
     		& 0.095  & 0.914
     		& 0.100  & 0.902
                & 0.372  & 0.689
                & 0.121  & 0.864
     		\\

    Train U-Net with LoRA & 16
                & 0.166  & 0.746
     		& 0.085  & 0.931
     		& 0.087  & 0.927
                & 0.352  & 0.712
                & 0.104  & 0.901
     		\\
       
    Train U-Net with LoRA & 64
                & 0.138  & 0.817
     		& 0.077  & 0.944
     		& 0.079  & 0.940
                & 0.336  & 0.734
                & 0.089  & 0.930
     		\\
       
    Train U-Net with LoRA & 256
                & 0.133  & 0.827
     		& 0.069  & 0.952
     		& 0.073  & 0.947
                & 0.325  & 0.745  
                & 0.088  & 0.933
     		\\
       
    Train U-Net with LoRA & 1024
                & 0.125  & 0.849
     		& 0.067  & 0.955
     		& 0.074  & 0.947  
                & 0.324  & 0.747  
                & 0.084  & 0.939
     		\\

    \bottomrule
  \end{tabular}
  }
  \label{tab: ablation_freeze_finetune_lora}
\end{table*}

\begin{tcolorbox}[
colframe=black,
arc=4pt,
boxsep=1pt,
]
    \paragraph{\textbf{\textit{Finding} 4.}} Fine-tuning the denoiser %
    appears to be 
    preferable for achieving better results, compared to either merely utilizing its intermediate features or training %
    a 
    LoRA.
\end{tcolorbox}

\subsection{Is the Training Data Quality Essential?}

The quality of annotations in real datasets is often lower compared to synthetic datasets, where data is precisely rendered via simulators. In \cref{tab: ablation_data_quality}, we explore the impact of data quality on the fine-tuning process. We sample the same distribution of real data, consisting of 90\% from approximately 50K indoor images from the Taskonomy dataset \citep{zamir2018taskonomy} and 10\% from about 40K outdoor images from the Cityscapes dataset \citep{cordts2016cityscapes}. With lower annotation quality, the model achieves slightly worse performance. Also, the visualization in the supplementary material indicates that noisy data significantly influences detailed predictions in visual perception tasks.

\begin{table*}[t]
  \centering
  \caption{Investigations into the impact of training data quality on depth estimation.}
\resizebox{\linewidth}{!}{%
  \begin{tabular}{@{}r|c|lr|lr|lr|lr|lr@{}}
    \toprule
	
	\multirow{2}{*}{Data Quality} & \multirow{2}{*}{Datasets} & \multicolumn{2}{c|}{KITTI}  & \multicolumn{2}{c|}{NYU} & \multicolumn{2}{c|}{ScanNet} 
 & \multicolumn{2}{c|}{DIODE} & \multicolumn{2}{c}{ETH3D}\\
	
    \cline{3-12}
	
    & & AbsRel$\downarrow$ & $\delta_1$$\uparrow$ & AbsRel$\downarrow$ & $\delta_1$$\uparrow$ & 
    AbsRel$\downarrow$ & $\delta_1$$\uparrow$ & AbsRel$\downarrow$ & $\delta_1$$\uparrow$ & AbsRel$\downarrow$ & $\delta_1$$\uparrow$ \\

    \hline

    Synthetic Data & Hypersim (50K) + Virtual KITTI (40K)
                & 0.100  & 0.902  
     		& 0.053  & 0.966  
     		& 0.059  & 0.961  
                & 0.309  & 0.768  
                & 0.068  & 0.956  
     		\\
    Real Data & Taskonomy (50K) + Cityscapes (40K)
                & 0.123  & 0.857  
     		& 0.055  & 0.966
     		& 0.062  & 0.958
                & 0.293  & 0.762  
                & 0.074  & 0.947
     		\\

    \bottomrule
  \end{tabular}
  }
  \label{tab: ablation_data_quality}
\end{table*}

\begin{tcolorbox}[
colframe=black,
arc=4pt,
boxsep=1pt,
]
    \paragraph{\textbf{\textit{Finding} 5.}} Data quality affects the fine-grained details of dense predictions significantly.
\end{tcolorbox}

\subsection{Summary of the obervations}
Based on the preceding analysis, an effective approach to leveraging the prior knowledge of diffusion models is to use them as single-step deterministic perception estimators. This can be done with either a VAE decoder or a customized lightweight decoder. Additionally, employing pixel-specific customized losses can further enhance detail and overall performance. We compare our deterministic single-step perception method with previous multi-step paradigms in \cref{fig: pipeline}. In the following section, we extend these findings to a broader set of visual perception tasks, including surface normal estimation, semantic image segmentation, dichotomous image segmentation, and image matting.

\section{Experiments on Various Dense Visual Perceptual Tasks}

In this section, we empirically show the robust transfer ability of our GenPercept on diverse visual tasks. Unless specified otherwise, we freeze the VAE AutoEncoder and fine-tune the U-Net of Stable Diffusion v2.1 to estimate the ground-truth label latent for 30000 iterations, with a resolution of (768, 768), a batch size of 32, and a learning rate of 3e-5. Different customized loss functions are utilized to improve the performance further on dense visual perception tasks.

\subsection{Geometric Estimation}
\label{sec: geometric_estimation}

\red{For geometry evaluation, the ensemble size, inference resolution, valid evaluation depth range (specific for depth estimation), and evaluation average paradigm (average by pixels or average by the number of images) can be different for each method. To compare these approaches fairly, we follow the open-source evaluation code of Marigold \citep{ke2023repurposing} for depth and DSINE \citep{bae2024dsine} for surface normal, and evaluate the performance of partial existing SOTA methods with their officially released model weights. They are labeled with $^\dagger$ in the Table.}

\noindent\textbf{Monocular Depth Estimation.} The monocular depth estimation aims to predict the vertical distance between the observed object and the camera from an RGB image. The estimated depth is formulated as affine-invariant depth \citep{Wei2021CVPR_leres,Ranftl2020_midas,ranftl2021_dpt}, and should be recovered by performing least square regression with the ground truth. The evaluation is performed on five zero-shoft datasets including KITTI \citep{KITTI}, NYU \citep{NYUv2}, ScanNet \citep{ScanNet},  DIODE \citep{diode_dataset}, and ETH3D \citep{schops2017multiEth3d}. We compute the absolute relative error (AbsRel$\downarrow$) and percentage of accurate valid depth pixels ($\delta_1$$\uparrow$). Invalid regions are filtered out and the metrics are averaged on all images.

\begin{table}[t]
  \centering
  \caption{Quantitative comparison of affine-invariant depth estimation on five zero-shot datasets. \red{Part of the reported results ($^\dagger$) are evaluated following the evaluation protocol of Marigold by ourselves.}
  }
  
\resizebox{.99\linewidth}{!}{%
  \begin{tabular}{@{}r|c|lr|lr|lr|lr|lr@{}}
    \toprule
	
	\multirow{2}{*}{Method} & Training & \multicolumn{2}{c|}{KITTI}  & \multicolumn{2}{c|}{NYU} & \multicolumn{2}{c|}{ScanNet}
 & \multicolumn{2}{c|}{DIODE} & \multicolumn{2}{c}{ETH3D}\\
	
    \cline{3-12}
	
    & Samples &  AbsRel$\downarrow$ & $\delta_1$$\uparrow$ & AbsRel$\downarrow$ & $\delta_1$$\uparrow$ & AbsRel$\downarrow$ & $\delta_1$$\uparrow$ & AbsRel$\downarrow$ & $\delta_1$$\uparrow$ & AbsRel$\downarrow$ & $\delta_1$$\uparrow$ \\

    \hline
       
    MiDaS \citep{Ranftl2020_midas}  & 2M	    & 0.236  & 0.630
     		& 0.111	& 0.885
                & 0.121 & 0.846
     		& 0.332	& 0.715
                & 0.184  & 0.752
     		\\
       
    Omnidata \citep{eftekhar2021omnidata}  & 12.2M	& 0.149  & 0.835
     		& 0.074	& 0.945
                & 0.075 & 0.936
     		& 0.339	& 0.742
                & 0.166  & 0.778
     		\\
       
    DPT-large \citep{ranftl2021_dpt}  & 1.4M	& 0.100  & 0.901
     		& 0.098	& 0.903
                & 0.082 & 0.934
     		& 0.182	& 0.758
                & 0.078 & 0.946
     		\\

    \textcolor{red}{DepthAnything$^\dagger$ \citep{yang2024depth}}  & 63.5M	& 0.080  & 0.946
     		& 0.043	& 0.980
                & 0.043  & 0.981
     		& 0.261	& 0.759
                & 0.058  & 0.984
     		\\

    \textcolor{red}{DepthAnything v2$^\dagger$ \citep{yang2024depthv2}}  & 62.6M	& 0.080  & 0.943
     		& 0.043	& 0.979
                & 0.042  & 0.979
     		& 0.321	& 0.758
                & 0.066  & 0.983
     		\\

    \textcolor{red}{Metric3D v2$^\dagger$ \citep{hu2024metric3dv2}}  & 16M	& \textbf{0.052}  & \textbf{0.979}
     		& \textbf{0.039}	& \textbf{0.979}
                & \textbf{0.023}  & \textbf{0.989}
     		& \textbf{0.147}	& \textbf{0.892}
                & \textbf{0.040}  & \textbf{0.983}
     		\\
    
    \hline
    \hline

    DiverseDepth \citep{yin2020diversedepth}  & 320K 	& 0.190  & 0.704
     		& 0.117	& 0.875
                & 0.109 & 0.882
     		& 0.376	& 0.631
                & 0.228 & 0.694
     		\\
       
    LeReS \citep{Wei2021CVPR_leres}  & 354K	    & 0.149  & 0.784
     		& 0.090	& 0.916
                & 0.091 & 0.917
     		& 0.271	& 0.766
                & 0.171 & 0.777
     		\\
       
    HDN \citep{zhang2022hierarchical}  & 300K	    & 0.115  & 0.867
     		& 0.069	& 0.948
                & 0.080 & 0.939
     		& 0.246	& \textbf{0.780}
                & 0.121  & 0.833
     		\\

    \red{GeoWizard \citep{geowizard}}  & \red{280K} & \red{0.097}  & \red{0.921}
     		& \red{\textbf{0.052}}	& \red{0.966}
                & \red{0.061} & \red{0.953}
     		& \red{0.297}	& \red{0.792}
                & \red{\textbf{0.064}}  & \red{\textbf{0.961}}
     		\\

    \red{DepthFM \citep{gui2024depthfm}}  & \red{63K}	& \red{0.083}  & \red{0.934}
     		& \red{0.065}	& \red{0.956}
                &  \red{-} & \red{-}
     		& \red{\textbf{0.225}} & \red{\textbf{0.800}}
                & \red{-}  & \red{-}
     		\\
            
    \hline

    \textcolor{red}{Marigold$^\dagger$ \citep{ke2023repurposing}}  & 74K	& 0.099  & 0.916
     		& 0.055	& 0.964
                & 0.064  & 0.951
     		& 0.308	& 0.773
                & 0.065  & 0.960
     		\\

    \textcolor{red}{DMP Official$^\dagger$ \citep{lee2023exploiting}}  & \red{-}  & \red{0.240}  & \red{0.622}
     		& \red{0.109}	& \red{0.891}
                & \red{0.146}    & \red{0.814}
     		& \red{0.361} 	& \red{0.706}
                & \red{0.128}    &  \red{0.857}
     		\\

    \textcolor{red}{GeoWizard$^\dagger$ \citep{geowizard}}  & 280K & 0.129  & 0.851
     		& 0.059	& 0.959
                & 0.066  & 0.953
     		& 0.328	& 0.753
                & 0.077  & 0.940 
     		\\

    \textcolor{red}{DepthFM$^\dagger$ \citep{gui2024depthfm}}  & 63K	& 0.174  & 0.718
     		& 0.082	& 0.932
                & 0.095  & 0.903
     		& 0.334 	& 0.729
                & 0.101  & 0.902
     		\\

    \hline
    Our GenPercept (Depth)  & 90K	& 0.094  & 0.923
     		& \textbf{0.052}	& 0.966
                & \textbf{0.056}  & \textbf{0.965}
     		& 0.302	& 0.767
                & 0.066  & 0.957
     		\\

    Our GenPercept (Disparity)  & 90K	& 0.080  & 0.934
     		& 0.058	& \textbf{0.969}
                & 0.063 & 0.960
     		& 0.226	& 0.741 
                & 0.096  & 0.959
     		\\
       
    Our GenPercept (Disparity + DPT head)  & 90K & \textbf{0.078}  & \textbf{0.935}
     		& 0.059	& 0.967
                & 0.064  & 0.961
     		& 0.228	& 0.740
                & 0.094  & \textbf{0.961}
     		\\
     
    \bottomrule
  \end{tabular}
  }
  \label{tab: affine_inv_depth}
\end{table}

\begin{table*}[t!]
\footnotesize
\setlength\tabcolsep{1.5pt}
\renewcommand{\arraystretch}{1.0}
\caption{
Quantitative comparison of surface normal estimation on three zero-shot datasets. We evaluate mean error$\downarrow$, median error $\downarrow$ (med.), and the percentages of pixels $\uparrow$ with five thresholds. \red{Part of the reported results ($^\dagger$) are evaluated following the evaluation protocol of DSINE by ourselves.}
}
\begin{center}
\resizebox{\columnwidth}{!}{%
\begin{tabular}{r|c|cc|ccccc|cc|ccccc|cc|ccccc}
\toprule
\multirow{2}{*}{Method} & Training
& \multicolumn{7}{c|}{NYU v2}
& \multicolumn{7}{c|}{ScanNet}
& \multicolumn{7}{c}{Sintel} \\
\cline{3-23}
& Samples & mean & med. & {\scriptsize $5.0^{\circ}$} & {\scriptsize $7.5^{\circ}$} & {\scriptsize $11.25^{\circ}$} & {\scriptsize $22.5^{\circ}$} & {\scriptsize $30^{\circ}$} 
& mean & med. & {\scriptsize $5.0^{\circ}$} & {\scriptsize $7.5^{\circ}$} & {\scriptsize $11.25^{\circ}$} & {\scriptsize $22.5^{\circ}$} & {\scriptsize $30^{\circ}$} 
& mean & med. & {\scriptsize $5.0^{\circ}$} & {\scriptsize $7.5^{\circ}$} & {\scriptsize $11.25^{\circ}$} & {\scriptsize $22.5^{\circ}$} & {\scriptsize $30^{\circ}$} \\
\hline
Omnidata v1~\citep{eftekhar2021omnidata} & 12.2M & 
23.1 & 12.9 & 21.6 & 33.4 & 45.8 & 66.3 & 73.6 &
22.9 & 12.3 & 21.5 & 34.5 & 47.4 & 66.1 & 73.2 &
41.5 & 35.7 & 3.0 & 5.8 & 11.4 & 30.4 & 42.0 \\
Ominidata v2~\citep{kar20223d} & 12.2M & 
17.2 & 9.7 & 25.3 & 40.2 & 55.5 & 76.5 & 83.0 &
16.2 & 8.5 & 29.1 & 44.9 & 60.2 & 79.5 & 84.7 &
40.5 & 35.1 & 4.6 & 7.9 & 14.7 & 33.0 & 43.5 \\

\textcolor{red}{Metric3D v2$^\dagger$~\citep{hu2024metric3dv2}} & 8.8M & 
\textbf{13.5} & \textbf{6.7} & \textbf{40.1} & \textbf{53.5} &  \textbf{65.9} & \textbf{82.6}  & \textbf{87.7} &
\textbf{11.8} & \textbf{5.5} & \textbf{46.6} & \textbf{60.7} &  \textbf{71.6}  & \textbf{85.4} & \textbf{89.7} & 
\textbf{22.8} & \textbf{14.2} & \textbf{18.4} & \textbf{28.5} & \textbf{41.6} & \textbf{66.7} & \textbf{75.8} \\

\hline
\hline

\textcolor{red}{Geowizard~\citep{geowizard}} & 280K & 
\red{17.0} & \red{-} & \red{-} & \red{-} & \red{56.5} & \red{-} & \red{-} & 
\red{15.4} & \red{-} & \red{-} & \red{-} & \red{61.6} & \red{-} & \red{-} & 
\red{-} & \red{-} & \red{-} & \red{-} & \red{-} & \red{-} & \red{-} \\

\textcolor{red}{DINSE$^\dagger$~\citep{bae2024dsine}} & 160K & 
\textbf{16.4} & 8.4 & 32.8 & 46.3 & 59.6 & 77.7 & 83.5 & 
16.2 & 8.3 & 29.8 & 45.9 & 61.0 & 78.7 & 84.4 & 
34.9 & 28.1 & \textbf{8.9} & \textbf{14.1} & \textbf{21.5} & 41.5 & 52.7 \\ 

\textcolor{red}{Geowizard$^\dagger$~\citep{geowizard}} & 280K & 
19.8 & 11.2 & 18.0 & 32.7 & 50.2 &73.0 & 79.9 & 
21.1 & 11.9 & 15.9 & 29.7 & 47.4& 70.7& 77.8& 
36.1 & 28.4 & 4.1 & 8.6 & 16.9 & 39.8 & 52.5 \\
\hline

Our GenPercept (Latent MSE loss) & 90K & 
17.4 & 9.5 & 23.3 & 40.0 & 56.3 & 76.8 & 83.0  &
16.3 & 8.9 & 25.8 & 42.7 & 59.6 & 79.4 & 84.8 &
44.4 & 31.6 & 3.4 & 7.5 & 15.0 & 37.0 & 48.0  \\

Our GenPercept (Image angular loss) & 90K & 
\textbf{16.4} & \textbf{8.0} & \textbf{33.3} & \textbf{47.8} & \textbf{60.9} & \textbf{78.3} & \textbf{83.7} &
\textbf{15.2} & \textbf{7.4} & \textbf{33.9} & \textbf{50.7} & \textbf{65.0} & \textbf{80.9} & \textbf{85.7} & 
\textbf{34.6} & \textbf{26.2} & 5.2 & 9.8 & 18.4 & \textbf{43.8} & \textbf{55.8} \\

\bottomrule
\end{tabular}
}
\end{center}
\label{table: normal_benchmark}
\end{table*}

\noindent\textbf{Surface Normal Estimation.}
The surface normal estimation aims to predict a vector perpendicular to tangent plane of the surface at each point P, which represents the orientation of the object's surface. For evaluation, we compute the angular error on three zero-shot datasets including NYU \citep{NYUv2}, ScanNet \citep{ScanNet}, and Sintel \citep{sintel}. The mean $\downarrow$, median $\downarrow$, and the percentages of pixels $\uparrow$ with error below thresholds [5$^{\circ}$, 7.5$^{\circ}$, 11.25$^{\circ}$, 22.5$^{\circ}$, 30$^{\circ}$] are reported. Invalid regions are filtered out and the metrics are averaged on all images.

\noindent\textbf{Quantitative Evaluation.} Quantitative results on monocular depth estimation and surface normal estimation are shown in \cref{tab: affine_inv_depth} and \cref{table: normal_benchmark}, respectively. 
Even trained on limited synthetic datasets only, our GenPercept shows much robustness and achieves promising performance on diverse unseen scenes. For monocular depth models, we train them with pixel-wise MSE loss, scale-shift-invariant loss \citep{Ranftl2020_midas}, and gradient loss \citep{Ranftl2020_midas}. Furthermore, our disparity model (inverse of the depth) shows much better performance on datasets with outdoor scenes, such as KITTI and DIODE, but less performance on indoor datasets. \red{Therefore, we suggest adopting the depth model for indoor scenes and the disparity model for outdoor scenes experimentally.} By replacing the VAE decoder with a lightweight DPT head \citep{ranftl2021_dpt}, GenPercept can infer faster without bearing the performance penalty. For surface normal estimation, the image angular loss brings significant performance improvement thanks to our one-step estimation paradigm.

\begin{figure*}[t]
    \centering
    \includegraphics[width=.95\linewidth]{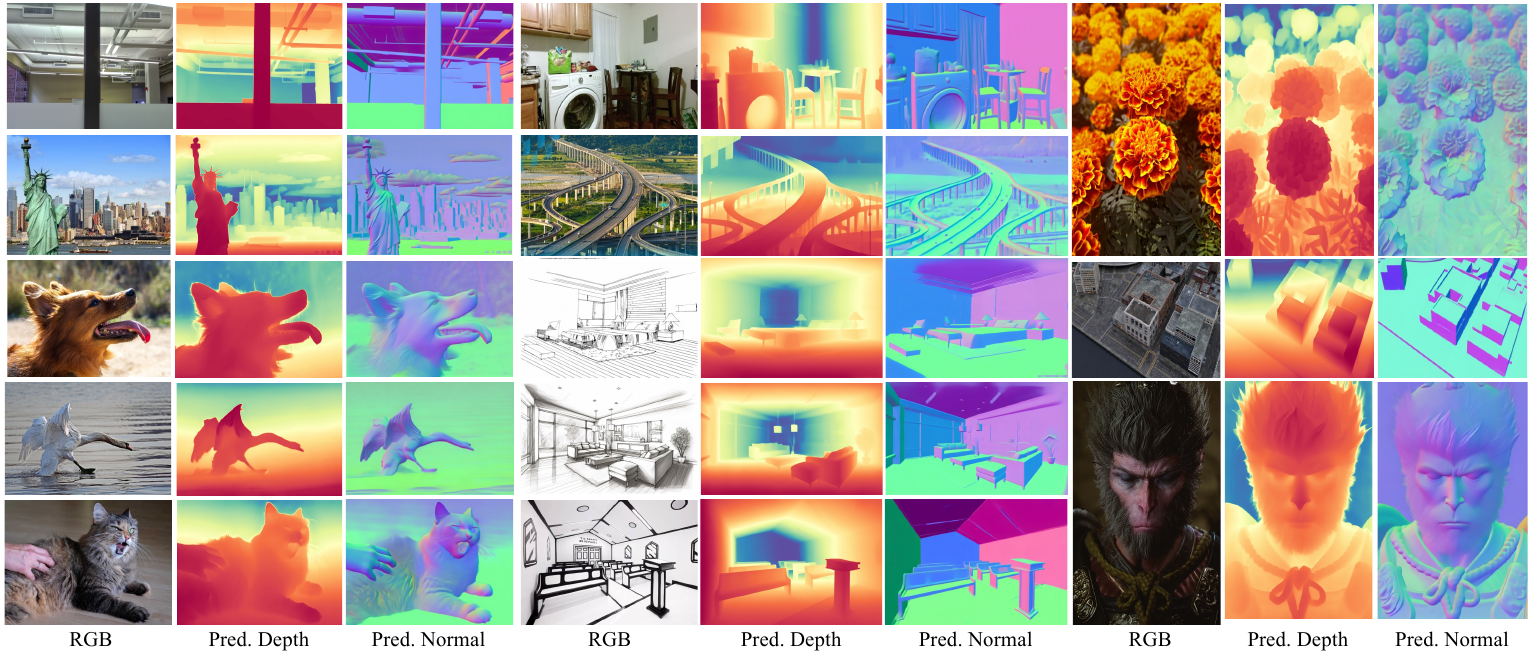}
    \caption{Qualitative results of monocular depth and surface normal estimation. The model works surprisingly well on \textit{out-of-domain images (sketch and cartoon images).}}
    \label{fig:rgb_depth_normal}
\end{figure*}

\noindent\textbf{Qualitative Results.} Qualitative visualizations are shown in \cref{fig:rgb_depth_normal}. %
{We observe excellent generalization of our models in that they can estimate accurate geometric information and promising details not only on diverse real and synthetic scenes, but also on comics, color drafts, and even sketches.}

\def\HCE{{\rm HCE}}
\subsection{Image Segmentation}
\label{sec: segmentation}

\begin{table}[b]
    \centering
    \caption{Quantitative results of dichotomous image segmentation on DIS5K validation and testing sets.
    Additional 
    cross-dataset evaluation is provided in the supplementary material.}
    \resizebox{\textwidth}{!}{
    \begin{tabular}{r|cccccc|cccccc|cccccc}
    \toprule
    Dataset & \multicolumn{6}{c|}{DIS-VD} & \multicolumn{6}{c|}{DIS-TE4} & \multicolumn{6}{c}{Overall DIS-TE (1-4)}\\
    \hline
    Metric & $maxF_\beta\uparrow$ & $F^w_\beta\uparrow$ & $~M\downarrow$ & $S_\alpha\uparrow$ & $E_\phi^m\uparrow$ & $\HCE_\gamma\downarrow$ & $maxF_\beta\uparrow$ & $F^w_\beta\uparrow$ & $~M\downarrow$ & $S_\alpha\uparrow$ & $E_\phi^m\uparrow$ & $\HCE_\gamma\downarrow$ & $maxF_\beta\uparrow$ & $F^w_\beta\uparrow$ & $~M\downarrow$ & $S_\alpha\uparrow$ & $E_\phi^m\uparrow$ & $\HCE_\gamma\downarrow$ \\
    \hline
    U$^2$Net~\citep{qin2020u2} & 0.748 & 0.656 & 0.090 & 0.781 & 0.823 & 1413 & 0.795 & 0.705 & 0.087 & 0.807 & 0.847 & 3653 & 0.761 & 0.670 & 0.083 & 0.791 & 0.835 & 1333 \\
    SINetV2~\citep{fan2021concealed} & 0.665 & 0.584 & 0.110 & 0.727 & 0.798 & 1568 & 0.699 & 0.616 & 0.113 & 0.744 & 0.824 & 3683 & 0.693 & 0.608 & 0.101 & 0.747 & 0.822 & 1411 \\
    HySM~\citep{nirkin2021hyperseg} & 0.734 & 0.640 & 0.096 & 0.773 & 0.814 & 1324 & 0.782 & 0.693 & 0.091 & 0.802 & 0.842 & 3331 & 0.757 & 0.665 & 0.084 & 0.792 & 0.834 & 1218 \\
    IS-Net~\citep{qin2022highly} & 0.791 & 0.717 & 0.074 & 0.813 & 0.856 & 1116 & 0.827 & 0.753 & 0.072 & 0.830 & 0.870 & 2888 & 0.799 & 0.726 & 0.070 & 0.819 & 0.858 & 1016 \\
    \red{MVANet~\citep{yu2024multi}} & \red{0.904} & \red{0.861} & \red{0.035} & \red{0.909} & \red{0.937} & \red{878} & \red{0.911} & \red{0.857} & \red{0.041} & \red{0.903} & \red{0.944} & \red{2301} & \red{0.916} & \red{0.855} & \red{0.035} & \red{0.905} & \red{0.938} & \red{790} \\
    \hline

    Our Genpercept & 0.857 & 0.835 & 0.04 & 0.87 & 0.934 & 1511 &
    0.848 & 0.829 & 0.049 & 0.854 & 0.938 & 3799 & 0.863 & 0.839 & 0.039 & 0.872 & 0.936 & 1364 \\

    \red{Our Genpercept (infer. at 1024px)} & \red{0.877} & \red{0.859} & \red{0.035} & \red{0.887} & \red{0.941} & \red{1262} & \red{0.874} & \red{0.858} & \red{0.041} & \red{0.874} & \red{0.947} & \red{3321} & \red{0.875} & \red{0.856} & \red{0.036} & \red{0.885} & \red{0.939} & \red{1176} \\
    
    \bottomrule
    \end{tabular}
    }
\label{tab:dis_mini}
\end{table}

\begin{table}[b]
  \centering
    \begin{minipage}[t]{0.45\linewidth}  %
    \centering
    \caption{
    Quantitative results of semantic segmentation on Hypersim and ADE20k.
    }
    \resizebox{.99\textwidth}{!}{
    \begin{tabular}{r|c|cc}
        \toprule
        \multirow{2}{*}{Method} & Training & mIoU$\uparrow$  & mIoU$\uparrow$  \\
        & Dataset & (Hypersim) & (ADE20K) \\
        \hline 
        GenPercept (Train UperNet) & \multirow{2}{*}{Hypersim}  & 46.0 & 34.1 \\
        GenPercept (Train U-Net + UperNet) &  & 52.9 & 38.3  \\
        \hline
        \red{Mask2Former R50} & \red{\multirow{4}{*}{ADE20K}} & \red{-} & \red{47.2}  \\
        \red{Mask2Former Swin-T} &  & \red{-} & \red{47.7}  \\
        \red{Mask2Former Swin-L} &  & \red{-} & \red{56.4}  \\
        \red{GenPercept (Train U-Net + UperNet)} &   & \red{-} & \red{50.2} \\
        \bottomrule
    \end{tabular}
    }
    \label{tab: semseg}
  \end{minipage}
  \hfill
  \begin{minipage}[t]{0.49\linewidth} %
    \centering
    \caption{Quantitative comparisons of image matting on the P3M-500-NP and AIM500.}
    \resizebox{\textwidth}{!}{
    \begin{tabular}{r|c|cccc}

    \toprule
    Method & Test Dataset& SAD $\downarrow$   & MAD $\downarrow$   & MSE $\downarrow$    & CONN $\downarrow$ \\ \hline
    HATT~\citep{Qiao_2020_CVPR} & \multirow{6}{*}{P3M-500-NP} & {30.35} & {0.0176} & {0.0072} & 27.42 \\
    SHM~\citep{chen2018semantichumanmatting} &  & {20.77} & {0.0122} & {0.0093} & 17.09 \\
    MODNet~\citep{MODNet} &  & {16.70} & {0.0097} & {0.0051} & 13.81 \\
    P3M-Net~\citep{li2021privacy} &  & 11.23 & 0.0065 & 0.0035 & 12.51 \\
    \red{ViTAE-S~\citep{ma2023rethinking}} &  & \red{\textbf{7.59}}  &  \red{\textbf{0.0044}} &  \red{\textbf{0.0019}} &  \red{\textbf{6.96}} \\
    Our GenPercept & & 12.77 & 0.0074 & 0.0027 & 10.46 \\ 
    \hline
    \red{ViTAE-S~\citep{ma2023rethinking}} & \red{AIM500} & \red{112.52}  &  \red{0.0608} &  \red{0.0602} &  \red{43.18} \\
    \red{Our GenPercept} & \red{(Zero-shot)} &  \red{\textbf{75.5}} & \red{\textbf{0.0444}} & \red{\textbf{0.0242}} & \red{\textbf{36.74}} \\

    \bottomrule
    
    \end{tabular}
    }
    \label{tab: matting}
    \end{minipage}
    \hfill
\end{table}

\noindent\textbf{Dichotomous Image Segmentation}. This is a category-agnostic, high-quality object segmentation task that accurately separates the object from the background in an image. 
Consistent with previous methods, we use the six evaluation metrics specified in the DIS task, which include maximal F-measure ($maxF_\beta\uparrow$)~\citep{achanta2009frequency}, weighted F-measure ($F^w_\beta\uparrow$)~\citep{margolin2014evaluate}, mean absolute error ($M\downarrow$)~\citep{perazzi2012saliency}, structural measure ($S_\alpha\uparrow$)~\citep{fan2017new}, mean enhanced alignment measure~($E_\phi^m\uparrow$)~\citep{fan2018enhanced,fan2021cognitive} and human correction efforts ($\HCE_\gamma\downarrow$)~\citep{qin2022highly}.
We choose DIS5K~\citep{qin2022highly} as the training and testing dataset. 
We utilize DIS-TR for training and evaluate our model on DIS-VD and DIS-TE subsets. The pixel-wise MSE loss is utilized during training. 

Quantitative results of dichotomous image segmentation are shown in \cref{tab:dis_mini}. We only show partial results due to paper page limitations, full comparisons are accessible in the supplementary material. \red{GenPercept outperforms methods like HySM \citep{nirkin2021hyperseg} and IS-Net \citep{qin2022highly} on this challenging dataset across most evaluation metrics, but there exists room for further improvement compared to SoTA methods like MVANet \citep{yu2024multi}. As shown in Fig.~\ref{fig:dis_comp}, our approach provides a detailed foreground mask. For thin lines and meticulous objects that are difficult for previous methods to process, our method can also output accurate segmentation results.}

\begin{figure}[t]
    \centering
    \includegraphics[width=1.0\linewidth]{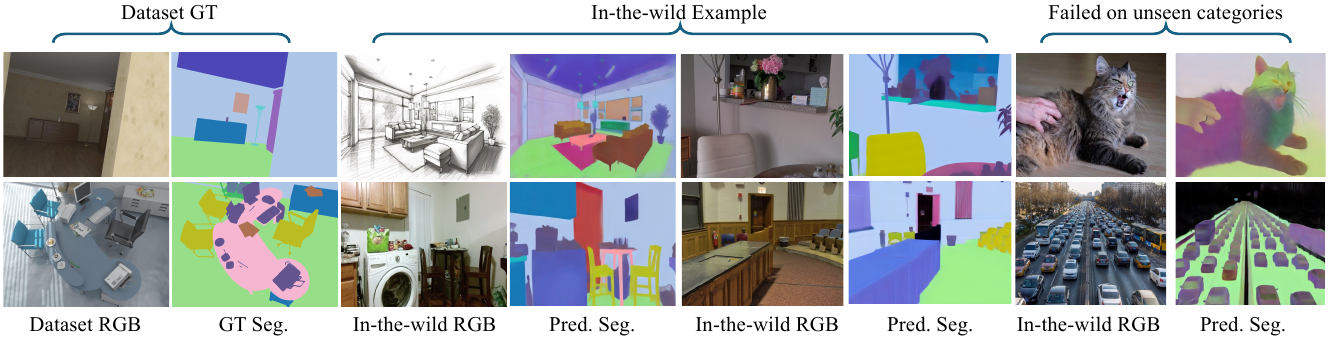}
    \caption{Quanlitative results of semantic segmentation in the wild. Trained on the synthetic indoor Hypersim dataset, GenPercept shows much robustness on the trained categories of complex in-the-wild images, \textit{e.g.}, yellow chairs, green floor, and light blue wall. Due to the limited annotation categories and little negative label of ``unknown category'', it sometimes fails in outdoor scenes and unseen categories such as cats and cars.}
    \label{fig:semseg_mask_vis}
\end{figure}

\begin{figure}[t]
    \centering
    \includegraphics[width=\linewidth]{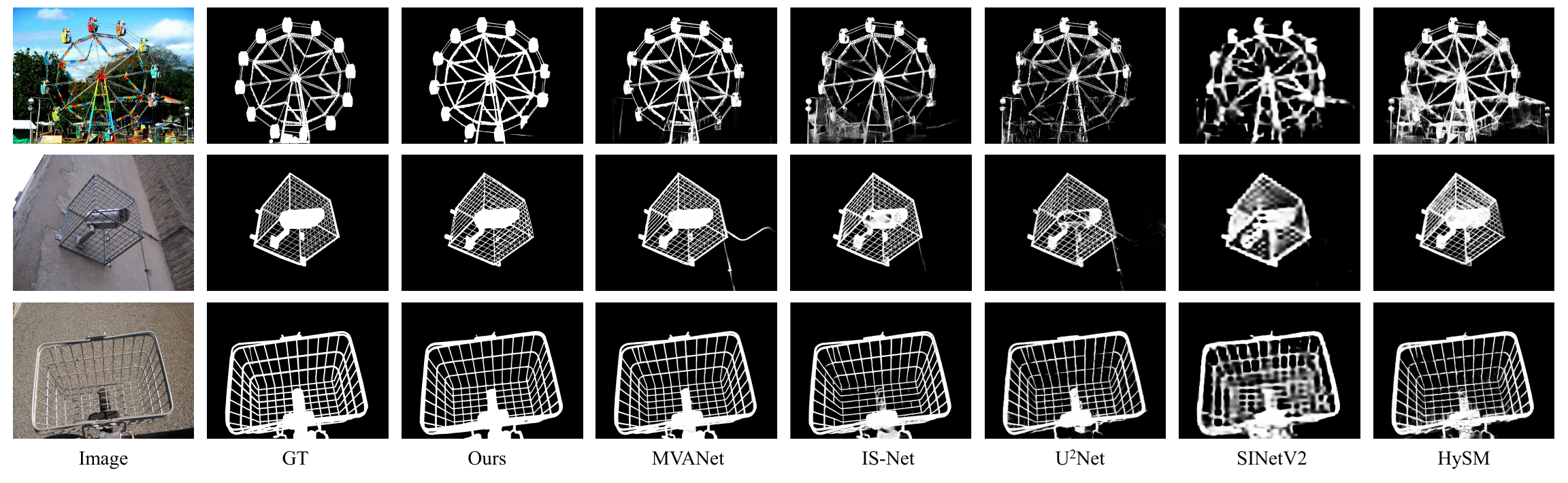}
    \caption{\red{Qualitative comparison of dichotomous image segmentation.}}
    \label{fig:dis_comp}
\end{figure}

\noindent\textbf{Semantic Image Segmentation.} This is a fundamental computer vision task that involves classifying each pixel in an image into a specific category or class. For training, we utilized the indoor synthetic dataset, HyperSim \citep{hypersim}, which comprises 40 semantic segmentation class labels. We encode different classes into 3-channel colormaps, treat the task as a regression problem, and fine-tune the original Stable Diffusion with the pixel-wise MSE loss. As demonstrated in \cref{fig:semseg_mask_vis}, the model generalizes well to classes within the HyperSim annotations, such as chairs and desks, but struggles with unrecognized categories such as cats and cars.

Another choice involves using a customized segmentation head. We incorporate a custom segmentation head, namely UperNet \citep{upernet}, onto the multi-level features extracted by UNet. \red{For the UperNet segmentation head, we follow the traditional semantic segmentation format to use n-channel output, where n is the number of categories.} The quantitative results are presented in \cref{tab: semseg}, we test the model's performance on Hypersim \citep{hypersim} and zero-shot ability on a subset of the ADE20k \citep{ade20k} validation set, which contains overlapping classes. \red{Besides, we compare with Mask2Former \citep{cheng2022masked} by training on ADE20K. GenPercept outperforms ResNet50 \citep{he2016deep} and Swin-T \citep{liu2021swin} of Mask2Former but achieves lower performance than Swin-L \citep{liu2021swin}.}

\subsection{Image Matting}
\label{sec: image_matting}

\noindent\textbf{Task Definition.}
Image matting aims to extract the foreground, background, and alpha mask from an image. Traditional approaches depend on supplementary inputs that delineate foreground, background, and ambiguous areas to reduce uncertainty. Automatic image matting seeks to remove this dependency by directly estimating these components from the image alone. The implementation details can be found in the supplementary material.

\noindent\textbf{Quantitative and Qualitative Results.} %
We evaluate metrics including the sum of absolute differences (SAD), mean squared error (MSE), mean absolute difference (MAD), gradient (Grad.), and Connectivity (Conn.) on the P3M-500-NP test set. SAD and MAD measure the mean L1 distance between predictions and ground truth labels. MSE and CONN focus on L2 distance and connectivity that better reflects human intuition. \red{As shown in \cref{tab: matting}, our GenPercept is less accurate compared with the state-of-the-art methods. However, when transferring the human image matting ability to general image matting tasks, GenPercept achieves much better performance. It proves the robustness brought by the prior knowledge of diffusion models pre-trained on the LAION dataset. Quantitative results of image matting are shown in \cref{fig: matting}. Please see supplementary for more visualization.}

\begin{figure}[t]
    \centering
    \includegraphics[width=.99\linewidth]{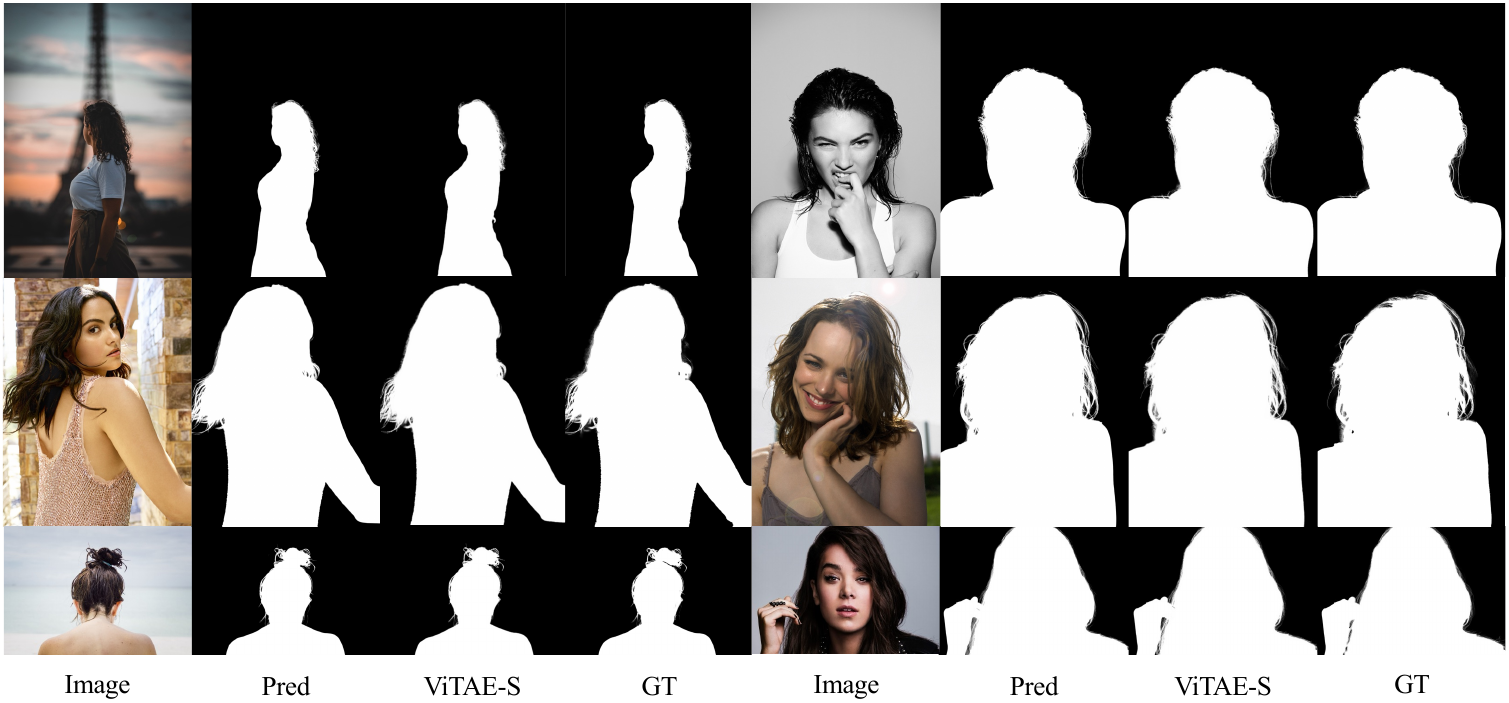}
    \caption{\red{Visualization of image matting on the P3M-500-NP test set.}}
    \label{fig: matting}
\end{figure}

\section{Related work}

\noindent\textbf{Vision Pre-Training.}
Models pretrained on large-scale datasets possess powerful feature extraction capabilities, enabling them to be effectively transferred to a wide range of visual tasks. For instance, the ResNet \citep{he2016deep} model pretrained on ImageNet~\citep{russakovsky2015imagenet} can be fine-tuned and applied to perception tasks. By means of contrastive learning, MoCo \citep{he2020momentum} and CLIP \citep{radford2021learning} acquire rich visual and semantic representations, leveraging their advantages in joint visual and semantic modeling to enhance the performance of multimodal tasks. DINO \citep{caron2021emerging}, through self-distillation, endows Vision Transformer and convolutional networks with comparable visual representation quality and demonstrates that self-supervised ViT representations contain explicit semantic segmentation information. \red{DINOv2 \citep{DINOV2} leverages self-supervised learning on a large curated dataset and exhibits remarkable zero-shot generalization capabilities across computer vision tasks at both image and pixel levels, including classification, semantic segmentation, and depth estimation.} In our work, we leverage Stable Diffusion \citep{Rombach_2022_CVPR} as a prior for scene understanding and transfer it to various perception tasks.

\noindent\textbf{Diffusion Priors for Dense Prediction.}
Several works explore to use the priors of generative models for perceptual tasks. Some works~\citep{bhattad2024stylegan, du2023generative} demonstrate that generative models encode property maps of the scene. By finding latent variable offsets, using LoRA~\citep{hu2022lora}, etc., generative models can directly produce intrinsic images like surface normals, depth, albedo, etc. 
LDMSeg~\citep{van2024simple} devises an image-conditioned sampling process, enabling diffusion models to directly output panoptic segmentation. UniGS~\citep{qi2023unigs} proposes location-aware color encoding and decoding strategies, allowing diffusion models to support referring segmentation and entity segmentation. Marigold~\citep{ke2023repurposing} fine-tunes diffusion model on limited synthetic data, enabling it to support affine-invariant monocular depth estimation and exhibit strong generalization performance. However, Marigold is time-consuming due to the need for multiple iterations of denoising. Additionally, the Gaussian noise leads to inconsistent results across inferences, requiring aggregation over multiple inferences.
\citet{Xiang2023ICCV} train a denoising auto-encoder for image classification. The difference of their method compared with traditional denoising auto-encoder is that input images are encoded into a latent code and denoising is performed in the latent space rather than the pixel space. They show good results on very small-scale datasets (CIFAR and ImageNet-tiny) to prove the concept and no results were reported on larger datasets. Furthermore, GeoWizard~\citep{fu2024geowizard} extends the generative capabilities of Marigold, achieving better performance in joint depth and normal estimation, which enhances applications like 3D reconstruction and novel view synthesis. Moreover, DepthFM~\citep{gui2024depthfm} addresses the speed challenge of Marigold by employing flow matching, offering a fast and efficient monocular depth estimation model.

\section{Conclusion}
In this work, we introduce GenPercept, an embarrassingly straightforward yet powerful approach to re-use the off-the-shelf UNet trained using diffusion processes. GenPercept demonstrates the capability to effectively leverage pre-trained diffusion models across a range of downstream dense perception tasks. We contend that our proposed methodology provides an efficient and potent paradigm for harnessing the capabilities of pre-trained diffusion models in dense visual perception tasks. 

{For future research, we suggest investigating the impact of scaling up the volume of fine-tuning data and exploring the key components of pre-training by applying alternative self-supervised pre-training methods on the LAION dataset, such as Masked Autoencoders (MAE) or Contrastive Language-Image Pretraining (CLIP). It will be helpful to clarify whether the highly detailed visual predictions produced by existing diffusion models are primarily driven by the extensive LAION dataset or the diffusion pretraining paradigm itself.}

 \section*{Acknowledgments}

 {

 
 Y. Ge is with The University of Adelaide and his 
 contribution was made when visiting Zhejiang University. 
 C. Shen is the corresponding author. 
 }

\clearpage

\definecolor{mygray}{rgb}{0.4858,0.3858,0.3858}
\newcommand{\csgray}[1]{{\color{mygray}{#1}}}

\appendix

\section{The formulation of Three Different Pipelines}

\subsection{Stochastic Multi-Step Generation}
\label{sec: stochastic_multi_Sstep_denoising}

For the training process, the RGB image $\image$ and ground-truth label $\gt$ are encoded into the latent space with the VAE encoder $\latentimage = \encoder(\image)$, $\latentgt = \encoder(\gt)$. 
The Gaussian noise $\noise$ is added to the ground-truth label latent $\latentgt$, and the noisy label latent $\latentgt_t$ is concatenated with the clean image latent $\latentimage$ as U-Net input
$\latent_t$ for each timestep:
\begin{equation}
\label{eq: multi_step_noise_blending}
\begin{gathered}
    \latentgt_t = \sqrt{\bar{\alpha}_t} \latentgt + \sqrt{1 - \bar{\alpha}_t} \noise, ~~t = [1, ..., T], \\
    \latent_t = \text{concat}(\latentgt_t, \latentimage),
\end{gathered}
\end{equation}
where $\bar{\alpha}_t = 
\prod_{s=1}^{t}{(1 - \beta_s)}$, and $\beta_s$ is sampled from a variance scheduler $\{ \beta_t \in (0, 1)\}_{t=1}^T$. \red{The scheduler is parameters with two hyperparameters $\beta_{start}$ and $\beta_{end}$, which defines the $\beta_t$ values of t=0 and t=1000, respectively. For a casual timestep $s$, $\beta_s$ is computed by linearly interpolating between $\sqrt{\beta_{start}}$ and $\sqrt{\beta_{end}}$, then squaring each interpolated value. The formulation is as follows.}
\begin{equation}
\begin{gathered}
    \red{\beta_s = (\sqrt{\beta_{start}} + \frac{s}{T}(\sqrt{\beta_{end}} - \sqrt{\beta_{start}}))^2, ~~s = [1, ..., T]}
\end{gathered}
\end{equation}
Then, the denoiser $\denoiser(\cdot, \cdot)$ is enforced to learn the ``v-prediction'' \citep{salimans2021progressive} from a timestep $t$ and the corresponding input $\latent_t$. During training, the parameters of VAE are frozen, and only the denoiser $\denoiser$ is fine-tuned. The timestep is uniformly sampled from 1 to $T$. 
\begin{equation}
\label{eq: multi_step_noise_loss}
\begin{gathered}
    \mathcal{L} = \mathbb{E}_{\latentgt, \noise \sim \mathcal{N}(0,I),t \sim \mathcal{U}(T)} \left\| (\sqrt{\bar{\alpha}_t} \noise - \sqrt{1 - \bar{\alpha}_t} \latentgt ) - \denoiser(\latent_t, t) \right\|^2_2.
\end{gathered}
\end{equation}

For inference, a Gaussian noise $\noise_t$ is randomly sampled and denoised step by step with the denoiser $\denoiser(\cdot, \cdot)$. 
\begin{equation}
\label{eq: multi_step_noise_inference}
\begin{gathered}
    \latenthatgt_T = \noise, ~~\latent_t = \text{concat}(\latenthatgt_t, \latentimage), ~~t = [T, ..., 1], \\
    \latenthatgt_{t\rightarrow 0} = \sqrt{\bar{\alpha}_t} \cdot \latenthatgt_t - \sqrt{1 - \bar{\alpha}_t} \cdot \denoiser(\latent_t, t), ~~
    \prednoise_t = \sqrt{\bar{\alpha}_t} \cdot \denoiser(\latent_t, t) - \sqrt{1 - \bar{\alpha}_t} \cdot \latenthatgt_t, \\
    \latenthatgt_{t-1} = \sqrt{\bar{\alpha}_{t-1}} \cdot \latenthatgt_{t\rightarrow 0} + \sqrt{1 - \bar{\alpha}_{t-1}} \cdot \prednoise_t , ~~
    \hat{\gt} = \decoder(\latenthatgt_{1\rightarrow0}).
\end{gathered}
\end{equation}
where the denoising process first computes the estimated noise $\prednoise_t$ and the predicted clean latent code $\latenthatgt_{t\rightarrow 0}$ of timestep $t$ from the current latent code $\latenthatgt_t$ and the predicted velocity $\denoiser(\latent_t, t)$. Then, it adds the computed noise $\prednoise_t$ back to $\latenthatgt_{t\rightarrow 0}$ to get the latent code of timestep $t-1$. After repeating it for $T$ times, the predicted clean latent code $\latenthatgt_{1\rightarrow0}$ is computed and sent to the VAE decoder $\decoder$ and estimate the target label $\hat{\gt}$. 
During inference, $m$ randomly sampled noises are introduced to estimate $m$ different predictions, and they are averaged with an ensemble process \citep{ke2023repurposing} to reduce the randomness of prediction and improve performance.

\subsection{Deterministic Multi-Step Generation}
\label{sec: deterministic_multi_step_inference}

The inherent random nature of diffusion models makes them challenging to apply to perceptual tasks, which typically aim for accurate results. As a result, existing works have replaced the original Gaussian noise with the target image as RGB noise \citep{cold_diffusion, lee2023exploiting}.
Technically,
rather than 
introducing
the random Gaussian noise $\noise$ , 
we blend 
the ground-truth label latent $\latentgt = \encoder(\gt)$ with the RGB image latent $\latentimage = \encoder(\image)$,
which is formulated as: 
\begin{equation}
\label{eq: rgb_noise_blending}
\begin{gathered}
    \latent_t = \latentgt_t = \sqrt{\bar{\alpha}_t} \latentgt + \sqrt{1 - \bar{\alpha}_t} \latentimage, ~~t = [1, ..., T], \\
\end{gathered}
\end{equation}

\begin{table}[t]
  \centering
  \caption{\red{Runtime comparison of three diffusion for perception pipelines on an RTX 4090 GPU.}}
\resizebox{\linewidth}{!}{%
  \begin{tabular}{@{}r|c|c|c|c@{}}
    \toprule
	
	Experimental Setting & Ensemble & Denoise Steps & Inference Time & GPU Memory \\

    \hline

    Stochastic Multi-step Generation (w.~ensemble)  	& 10 & 10 & $\sim$5.74s & 16GB \\

    Stochastic Multi-step Generation (w/o ensemble)  	& 1 & 10 & $\sim$0.79s & 6.95GB
     		\\

    Deterministic Multi-step Generation  & 1 & 10 & $\sim$0.79s & 6.95GB
     		\\
    
    Deterministic One-step Inference  (Ours)  	& 1 & 1 & $\sim$0.34s & 6.95GB
     		\\
            
    Deterministic One-step Inference + DPT head (Ours)  	& 1 & 1 & $\sim$0.24s & 6.32GB
     		\\
    \hline

    \red{Metric3Dv2 \citep{hu2024metric3dv2}}  & \red{1}  & \red{1}  & \red{$\sim$0.25s}  & \red{2.63GB}  \\
    
    \red{DepthAnythingv2 \citep{yang2024depthv2}}  & \red{1}  & \red{1}  & \red{$\sim$0.07s}  & \red{2.82GB} \\

    \red{DSINE \citep{bae2024dsine}}  & \red{1}  & \red{1}  & \red{$\sim$0.18s}  & \red{2.23GB} \\

    \hline

    \red{Marigold \citep{ke2023repurposing}} & \red{1}  & \red{10} & \red{$\sim$0.79s}  & \red{6.95GB}   \\

    \red{GeoWizard \citep{geowizard}} & \red{1}  & \red{1}  & \red{$\sim$1.32s} & \red{6.81GB} \\

    \red{DepthFM \citep{gui2024depthfm}} & \red{1} & \red{2} & \red{$\sim$0.41s}  & \red{6.97GB}  \\

    \red{Our GenPercept (DPT head)} & \red{1} & \red{1}  & \red{$\sim$0.24s}  & \red{6.32GB} \\

    \bottomrule
  \end{tabular}
  }
  \label{tab: inference_time}
\end{table}

\begin{table*}[t]
  \centering
  \caption{\red{The impact of training data volume on affine-invariant monocular depth estimation.}}
\resizebox{\linewidth}{!}{%
  \begin{tabular}{@{}r|lr|lr|lr|lr|lr@{}}
    \toprule
	
	\multirow{2}{*}{Amount of Data} & \multicolumn{2}{c|}{KITTI}  & \multicolumn{2}{c|}{NYU} & \multicolumn{2}{c|}{ScanNet} 
 & \multicolumn{2}{c|}{DIODE} & \multicolumn{2}{c}{ETH3D}\\
	
    \cline{2-11}
	
    & AbsRel$\downarrow$ & $\delta_1$$\uparrow$ & AbsRel$\downarrow$ & $\delta_1$$\uparrow$ & AbsRel$\downarrow$ & $\delta_1$$\uparrow$ & AbsRel$\downarrow$ & $\delta_1$$\uparrow$ & AbsRel$\downarrow$ & $\delta_1$$\uparrow$ \\

    \hline
    
    90K (1/1)  	& 0.100  & 0.902
     		& 0.053	& 0.966
     		& 0.059  & 0.961
                & 0.309  & 0.768
                & 0.068  & 0.956
     		\\
    45K (1/2)  	& 0.101  & 0.902
     		& 0.056	& 0.964
     		& 0.058  & 0.963
                & 0.311  & 0.764
                & 0.070  & 0.955
     		\\
    22.5K (1/4)  	& 0.109  & 0.884
     		& 0.056	& 0.963
     		& 0.059  & 0.963
                & 0.322  & 0.754
                & 0.073  & 0.950
     		\\
    11.2K (1/8)  	& 0.117  & 0.866
     		& 0.060	& 0.962
     		& 0.065  & 0.957
                & 0.324  & 0.753
                & 0.076  & 0.943
     		\\
    5.6K (1/16)  	& 0.117  & 0.868
     		& 0.063	& 0.958
     		& 0.068  & 0.952
                & 0.331  & 0.744
                & 0.084  & 0.932
     		\\
    \bottomrule
  \end{tabular}
  }
  \label{tab: ablation_amount_data}
\end{table*}

Furthermore, the input latent has been modified to the latent code of input image $\latentimage$ instead of random Gaussian.
Consequently,
the learning objective function and the inference process are reformulated as follows. 
\begin{equation}
\label{eq: rgb_noise_loss}
\begin{gathered}
\mathcal{L} = \mathbb{E}_{(\latentimage, \latentgt), t \sim \mathcal{U}(T)} \left\| (\sqrt{\bar{\alpha}_t} \latentimage - \sqrt{1 - \bar{\alpha}_t} \latentgt ) - \denoiser(\latent_t, t) \right\|^2_2.
\end{gathered}
\end{equation}
\begin{equation}
\label{eq: rgb_noise_inference}
\begin{gathered}
    \latenthatgt_T = \latentimage, ~~t = [T, ..., 1], \\
    \latenthatgt_{t\rightarrow 0} = \sqrt{\bar{\alpha}_t} \cdot \latenthatgt_t - \sqrt{1 - \bar{\alpha}_t} \cdot \denoiser(\latenthatgt_t, t), 
    \latenthatimage_t = \sqrt{\bar{\alpha}_t} \cdot \denoiser(\latenthatgt_t, t) - \sqrt{1 - \bar{\alpha}_t} \cdot \latenthatgt_t, \\
    \latenthatgt_{t-1} = \sqrt{\bar{\alpha}_{t-1}} \cdot \latenthatgt_{t\rightarrow 0} + \sqrt{1 - \bar{\alpha}_{t-1}} \cdot \latenthatimage_t , ~~
    \hat{\gt} = \decoder(\latenthatgt_{1\rightarrow0}).
\end{gathered}
\end{equation}
where $\latenthatimage_t$ 
denotes the predicted RGB noise of timestep $t$.

\subsection{\Ours: Deterministic One-step Perception}
\label{sec: method_of_genpercept}

In the main paper, We set the ($\beta_{start}$, $\beta_{end}$) values to 1, and $\bar{\alpha}_t = 
\prod_{s=1}^{t}{(1 - \beta_s)} = 0$,  the formulation of \cref{eq: rgb_noise_blending} to \cref{eq: rgb_noise_inference} can be greatly simplified as follows.
\begin{equation}
\label{eq: one_step_blending}
\begin{gathered}
    \latent_t = \latentgt_t = \latentimage, ~~
\mathcal{L} = \mathbb{E}_{(\latentimage, \latentgt), t \sim \mathcal{U}(T)} \left\| - \latentgt - \denoiser(\latent_t, t) \right\|^2_2, \\
    \latenthatgt_T = \latentimage, ~~
    \latenthatgt_{t-1} = - \latenthatgt_t, ~~
    \latenthatgt_{1\rightarrow 0} = - \denoiser(\latenthatgt_1, t=1), ~~\hat{\gt} = \decoder(\latenthatgt_{1\rightarrow 0}).
\end{gathered}
\end{equation}

\noindent\textbf{One-step prediction.}
The input of the U-Net is an RGB latent code, and the output becomes the negative value of the ground-truth latent code, with no relationship to the timestep $t$. Therefore, we set the number of timesteps $T$ to 1 while preserving the same performance.
\begin{equation}
\label{eq: one_step_final_loss}
\begin{gathered}
\mathcal{L} = \mathbb{E}_{(\latentimage, \latentgt)} \left\| - \latentgt - \denoiser(\latentimage, t=1) \right\|^2_2, \\
    \latenthatgt_{1\rightarrow 0} = - \denoiser(\latentimage, t=1), ~~\hat{\gt} = \decoder(\latenthatgt_{1\rightarrow 0}).
\end{gathered}
\end{equation}

\section{A Difference Between Image Generation and Visual Perception}

\red{In text-guided image generation, a single textual input can correspond to an immense variety of potential images. This inherent uncertainty makes generating a high-quality image directly from random noise in a single step extremely challenging. Therefore, the multi-step generation enables the model to incrementally remove noise, progressively refining details and textures at each stage, which effectively simplifies the task. However, visual perception tasks conditioned on an RGB image are deterministic without any randomness, and such an easy injective mapping can be estimated with a one-step inference process, as most of the traditional visual perception methods do.

While Marigold series algorithms aim to leverage diffusion models' ability of generating highly detailed images to enhance visual perception with precise details, reformulating straightforward deterministic tasks as a denoising process can further simplify this problem, enforcing the network to exploit "shortcuts", as described in Section 3.1 of the main paper and illustrated in \cref{fig: viz_scheduler_supp}.}

\section{Runtime Analysis}

\red{In this section, we quantitatively analyze the inference times of the three aforementioned pipelines, as summarized in \cref{tab: inference_time}. The runtime is evaluated by averaging the inference times over 100 images with a resolution of $768\times768$, using an RTX 4090 GPU. For "Stochastic Multi-step Generation" methods \citep{ke2023repurposing, geowizard, gui2024depthfm}, such as Marigold \citep{ke2023repurposing}, they rely on an ensemble process where multiple inferences are performed with varying random noise inputs to mitigate uncertainties introduced by Gaussian noise. Consequently, this approach is computationally expensive. On the other hand, "Deterministic Multi-step Generation" methods \citep{lee2023exploiting} involve multiple denoising steps, which significantly reduce inference efficiency.

In contrast, our proposed one-step inference paradigm demonstrates a runtime that is 94\% and 57\% less than those of multi-step methods with ensemble and without ensemble, respectively. Furthermore, by incorporating a customized head, such as the DPT head \citep{ranftl2021_dpt}, both runtime and GPU memory requirements are further reduced by 27\% without compromising performance, maintaining a competitive level of accuracy and robustness.}

\red{Compared to existing state-of-the-art diffusion-based methods, our proposed GenPercept achieves a notable improvement in inference speed, attributed to the innovative one-step inference paradigm and the customized head. While our method demonstrates inference speeds comparable to Metric3Dv2 \citep{hu2024metric3dv2} and DSINE \citep{bae2024dsine}, it falls behind DepthAnythingV2 \citep{yang2024depthv2}. Note that the superior performance of DepthAnythingV2 is facilitated by its training on a relatively lightweight model, bolstered by extensive labeled and unlabeled datasets, and supported by substantial computational resources distributed across multiple GPUs.}

 \begin{figure}[t]
    \centering
    \includegraphics[width=\textwidth]{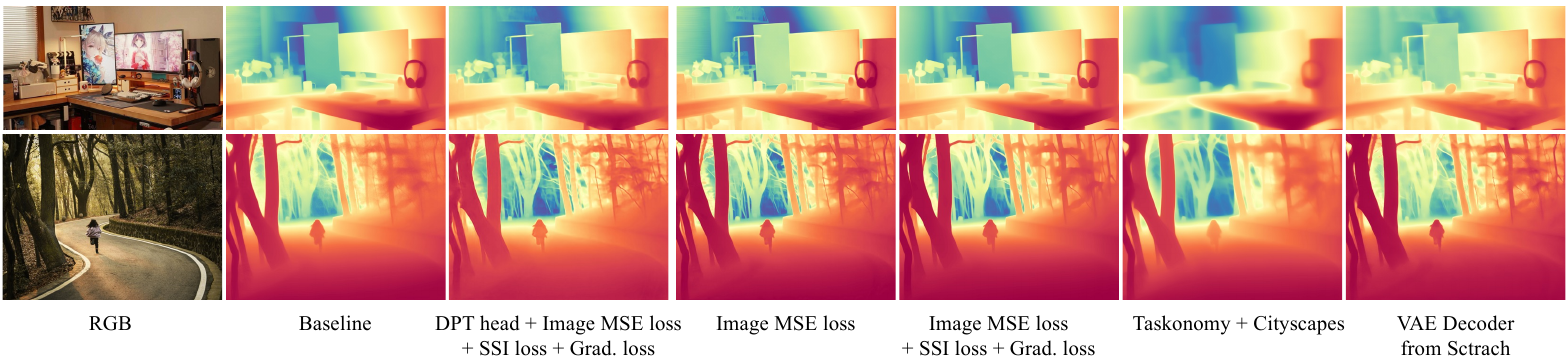}
    \vspace{-1 em}
    \caption{\red{Qualitative visualization of ablation study.}}
    \label{fig: viz_ablation}
\end{figure}

 \begin{figure}[t]
    \centering
    \includegraphics[width=\textwidth]{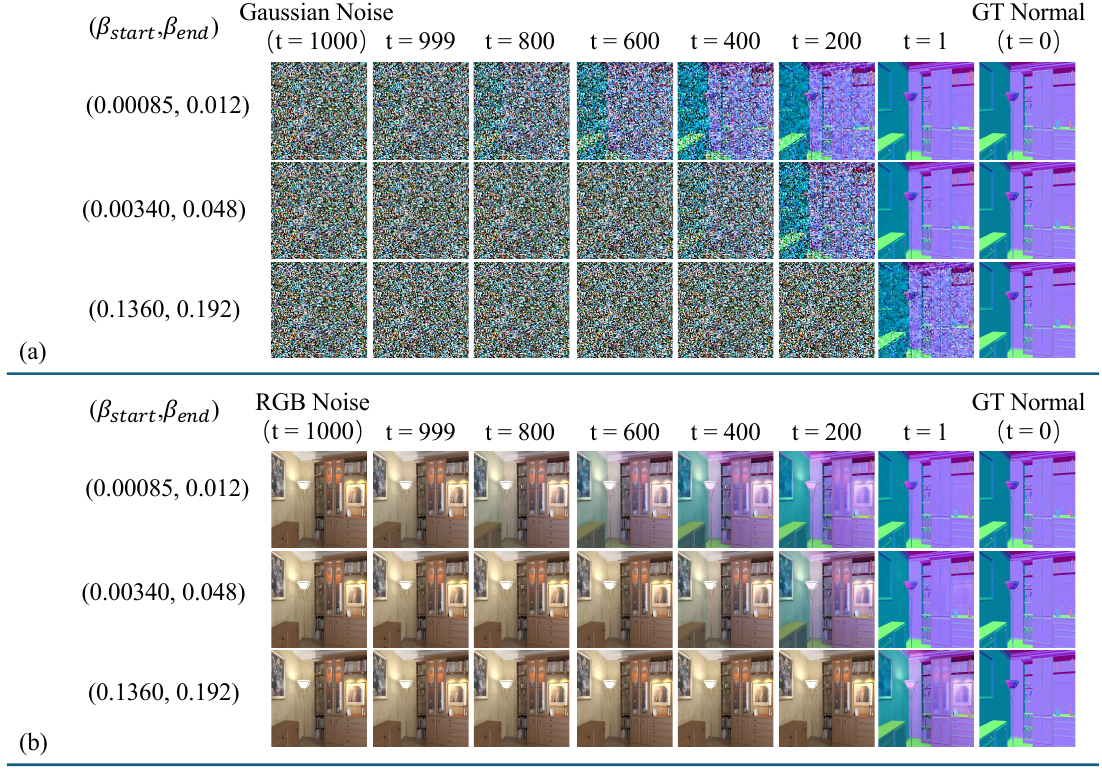}
    \vspace{-1 em}
    \caption{\red{Visualization of different noise forms and proportions in the forward diffusion process.}}
    \label{fig: viz_scheduler_supp}
\end{figure}

\section{The Impact of Data Volume}

\red{With only around 50K Hypersim \citep{hypersim} and 40K Virtual KITTI \citep{cabon2020vkitti2} fine-tuning data, GenPercept can generalize well to diverse tasks and datasets. How much data is needed for transferring at least? We gradually reduce the amount of training data. As shown in \cref{tab: ablation_amount_data}, less training data results in slightly worse performance, but GenPercept still shows much robustness to the data volume. }

\section{More Visualization Analysis}
\label{sec: more_vis_analysis}

\textbf{Quantitative Comparisons of Ablations.} Visualization of the ablation study experiments in the main paper \S 2 is shown in \cref{fig: viz_ablation}. The estimated depth detail remains comparable with a customized ``DPT head'' \citep{ranftl2021_dpt}. Models trained with lower-quality data like ``Taskonomy + Cityscapes'' or without the pre-trained VAE decoder parameters suffer from a decline in the ability to predict details. We attribute them to the low-quality annotation and the oversized decoder, respectively.

\textbf{Forward Diffusion Process.} More detailed visualization of different noise forms and proportions in the forward diffusion process is shown in \cref{fig: viz_scheduler_supp}. With larger ($\beta_{start}$, $\beta_{end}$) values of the DDIM scheduler, the proportion of noise will be much higher during the forward diffusion process. When ($\beta_{start}$, $\beta_{end}$) reaches (1.0, 1.0), the noisy latent will be pure noise latent, which is Gaussian noise and RGB latent for (a) and (b), respectively.

\begin{figure}[t]
    \centering
    \includegraphics[width=1\linewidth]{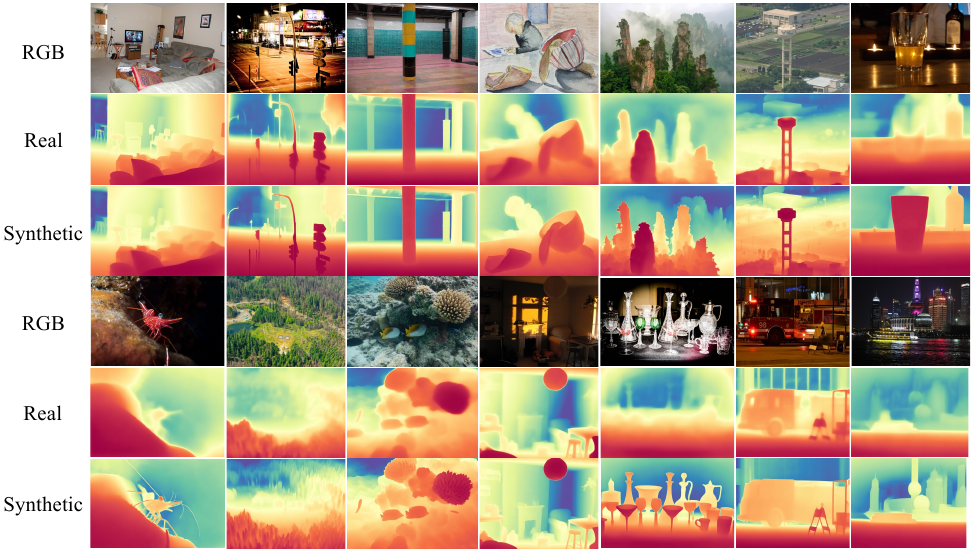}
    \caption{\red{Quantitative comparisons of generalization for models trained on synthetic and real data.}}
    \label{fig:real_sync_robust_compare}
\end{figure}

\red{\textbf{Quantitative Comparisons of Generalization.} We compare the generalization performance of models trained on synthetic (50K Hypersim \citep{hypersim} + 40K Virtual KITTI \citep{cabon2020vkitti2}) and real data (50K Taskonomy \citep{zamir2018taskonomy} + 40K Cityscapes \citep{cordts2016cityscapes}) for out-of-distribution scenarios. As illustrated in \cref{fig:real_sync_robust_compare}, the robustness of GenPercept trained on real data is comparable to that trained on synthetic data in challenging scenes, such as underwater environments, non-realistic renderings, and evening settings. Notably, models trained on synthetic data demonstrate superior accuracy in estimating transparent objects and capturing geometric details, owing to the high-quality and densely labeled synthetic ground truth.}

\section{An Extra Attempt on Human Pose Estimation}
\label{sec: human_pose}
\noindent\textbf{Task Definition}
Human Pose Estimation is a task aimed at determining the spatial configuration of a person or object in a given image or video. This involves identifying and predicting the coordinates of particular keypoints.

\begin{figure}[h]
    \centering
    \includegraphics[width=1\linewidth]{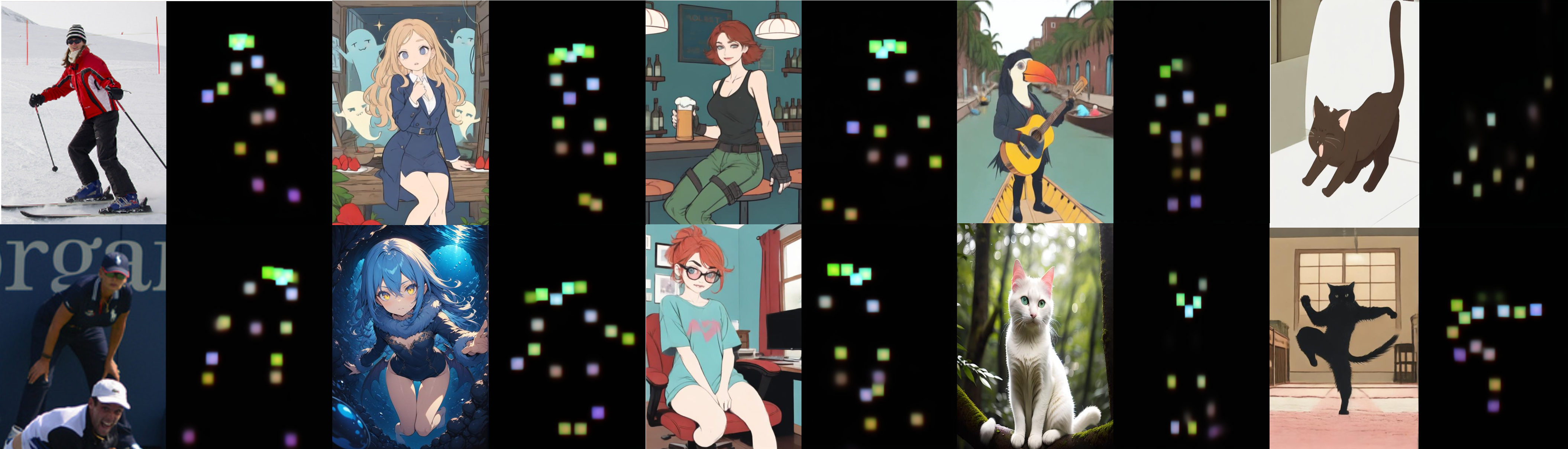}
    \caption{Generalized test results of keypoint detection.}
    \label{fig:keypoint}
\end{figure}

\noindent\textbf{Implementation Details}
For human keypoint detection, we use Simple Baseline \citep{xiao2018simple} for person detection and conduct training on the COCO training set with 15K training samples. Performance is evaluated on the COCO~\citep{lin2014microsoft} validation set. As shown in \cref{fig:keypoint}, it is generalizable to unseen objects in the training set. To evaluate the performance on COCO, we use the customized keypoint head of ViTPose~\citep{xu2022vitpose} for decoding the output. The quantitative results compared with a generalist model, Painter\citep{Painter} is shown in \cref{tab: keypoint}.

\begin{table*}[h]
  \centering
  \caption{Pose estimation on COCO.}
  \begin{tabular}{c|c|c|c|c|c}

    \toprule
    Metrics & AP $\uparrow$    & AP .5 $\uparrow$ & AP .75 $\uparrow$ & AP (M) $\uparrow$ & AP (L) $\uparrow$ \\
    \hline
    Painter \citep{Painter} & 0.721 & 0.900 & 0.781  & 0.686  & \textbf{0.786}  \\
    \hline
    {\Ours} &\textbf{0.752} &\textbf{0.907} & \textbf{0.824}  & \textbf{0.691}  & 0.778   \\
    \bottomrule
    
  \end{tabular}
  \label{tab: keypoint}
\end{table*}

\section{Implementation Details}

\noindent\textbf{\red{Implementation Details of Exploration Experiments in Section 3 of the Main Paper.}}
\red{Unless specified otherwise, we follow Marigold's training setting and train for 30000 iterations to estimate the affine-invariant monocular depth. The training dataset contains 50K Hypersim \citep{hypersim} images and 40K Virtual KITTI \citep{cabon2020vkitti2} images, and these images are sampled with a sampling rate of 90\% for Hypersim and 10\% for Virtual KITTI.} We freeze the VAE AutoEncoder and fine-tune the U-Net of Stable Diffusion v2.1 to estimate the ground-truth label latent, with a resolution of (480, 640), a batch size of 32, and a learning rate of 3e-5. The multi-resolution noise \citep{ke2023repurposing} is employed for Gaussian noise and not used for RGB noise \citep{lee2023exploiting} by default. For inference, the ensemble size and denoising steps are set to 1 and 10, respectively. Evaluation results of absolute relative error (AbsRel) and $\delta_1$ are reported on five unseen monocular depth datasets, including KITTI \citep{KITTI}, NYU \citep{NYUv2}, ScanNet \citep{ScanNet}, DIODE \citep{diode_dataset}, and ETH3D \citep{schops2017multiEth3d}.

\section{More Quantitative and Qualitive Experiments}

\subsection{Monocular Depth Estimation}

\noindent\textbf{More Qualitative Evaluation. }
We show the robustness of the monocular depth estimation model in diverse scenes in \cref{fig:supp_depth}. Compared to DPT \citep{ranftl2021_dpt}, our Genpercept performs better on estimating details and shows much better robustness even on some sketches. Compared to Marigold \citep{ke2023repurposing}, our method achieves better relative depth visualization.

\subsection{Surface Normal Estimation}
\noindent\textbf{More Qualitative Evaluation. } In \cref{fig:supp_normal}, we showcase more qualitative results for the surface normal estimation. DSINE \citep{bae2024rethinking} is trained on 160K images of 10 datasets, including both real and synthetic data. Our GenPercept is only trained on one synthetic dataset (Hypersim \citep{hypersim}), and can estimate much more detailed surface normal maps.

\subsection{Dichotomous Image Segmentation}
\def\HCE{{\rm HCE}}

\noindent\textbf{More Quantitative Evaluation.}
To conduct a comprehensive evaluation, we compare our approach with numerous previous models including models for medical image segmentation~\citep{ronneberger2015u}, salient object detection~\citep{qin2019basnet,zhao2020suppress,wei2020f3net,chen2020global,qin2020u2}, camouflaged object detection~\citep{fan2021concealed,mei2021camouflaged}, semantic segmentation~\citep{zhao2017pyramid,chen2018encoder,wang2020deep,yu2018bisenet,zhao2018icnet,howard2019searching,fan2021rethinking,nirkin2021hyperseg} and models like IS-Net~\citep{qin2022highly} \red{and MVANet \citep{yu2024multi}} specifically trained for DIS. 

\red{As shown in~\cref{tab:dis}, our proposed model significantly outperforms methods like IS-Net across most evaluation metrics on this challenging dataset. Compared with MVANet, which inferences with a resolution of $1024\times1024$, our GenPercept achieves slightly lower performance, but the results remains competitive and can highlight the effectiveness of our approach for DIS.}

\begin{table}[!ht]
    \centering
    \caption{Quantitative results on DIS5K validation and testing sets.}

    \resizebox{\textwidth}{!}{
    \begin{tabular}{c|c|cccccccccccccccccc|cc}
    \toprule
        Dataset & Metric & \tabincell{c}{UNet}& \tabincell{c}{BASNet} & \tabincell{c}{GateNet} & \tabincell{c}{F$^3$Net}&\tabincell{c}{GCPANet} & \tabincell{c}{U$^2$Net}&\tabincell{c}{SINetV2}&\tabincell{c}{PFNet}&\tabincell{c}{PSPNet}&\tabincell{c}{DLV3+}&
        \tabincell{c}{HRNet} &
        \tabincell{c}{BSV1}&\tabincell{c}{ICNet}&
        \tabincell{c}{MBV3} & \tabincell{c}{STDC}&\tabincell{c}{HySM}  & \tabincell{c}{IS-Net} & \red{\tabincell{c}{MVANet}} & \tabincell{c}{\textbf{Ours}} & \tabincell{c}{\textbf{Ours (1024px)}} \\
    \midrule
\multirow{6}*{DIS-VD} & $maxF_\beta\uparrow$   & 0.692 & 0.731  & 0.678   & 0.685    & 0.648   & 0.748    & 0.665   & 0.691 & 0.691  & 0.660 & 0.726 & 0.662 & 0.697 & 0.714 & 0.696 & 0.734 & 0.791 & \red{0.904} &  0.857 & \red{0.877} \\
~      & $F^w_\beta\uparrow$    & 0.586 & 0.641  & 0.574   & 0.595    & 0.542   & 0.656    & 0.584   & 0.604 & 0.603  & 0.568 & 0.641 & 0.548 & 0.609 & 0.642 & 0.613 & 0.640 & 0.717  & \red{0.861}  &  0.835 & \red{0.859} \\
~      & $~M~\downarrow$        & 0.113 & 0.094  & 0.110   & 0.107    & 0.118   & 0.090    & 0.110   & 0.106 & 0.102  & 0.114 & 0.095 & 0.116 & 0.102 & 0.092 & 0.103 & 0.096 & 0.074  &  \red{0.035}  &  0.040 & \red{0.035} \\
~      & $S_\alpha\uparrow$     & 0.745 & 0.768  & 0.723   & 0.733    & 0.718   & 0.781    & 0.727   & 0.740 & 0.744  & 0.716 & 0.767 & 0.728 & 0.747 & 0.758 & 0.740 & 0.773 & 0.813  &  \red{0.909}  &  0.870 & \red{0.887} \\
~      & $E_\phi^m\uparrow$     & 0.785 & 0.816  & 0.783   & 0.800    & 0.765   & 0.823    & 0.798   & 0.811 & 0.802  & 0.796 & 0.824 & 0.767 & 0.811 & 0.841 & 0.817 & 0.814 & 0.856  &  \red{0.937}  &  0.934 & \red{0.941} \\
~      & $\HCE_\gamma\downarrow$ & 1337  & 1402   & 1493    & 1567     & 1555    & 1413     & 1568    & 1606  & 1588   & 1520  & 1560  & 1660  & 1503  & 1625  & 1598  & 1324  & 1116  & \red{878}  &   1511 & \red{1262} \\
\midrule
\multirow{6}*{DIS-TE1} & $maxF_\beta\uparrow$   & 0.625 & 0.688  & 0.620   & 0.640    & 0.598   & 0.694    & 0.644   & 0.646 & 0.645  & 0.601 & 0.668 & 0.595 & 0.631 & 0.669 & 0.648 & 0.695 & 0.740 &  \red{0.893}  &  0.841 & \red{0.850} \\
~      & $F^w_\beta\uparrow$    & 0.514 & 0.595  & 0.517   & 0.549    & 0.495   & 0.601    & 0.558   & 0.552 & 0.557  & 0.506 & 0.579 & 0.474 & 0.535 & 0.595 & 0.562 & 0.597 & 0.662  &  \red{0.823}  &  0.814 & \red{0.827} \\
~      & $~M~\downarrow$        & 0.106 & 0.084  & 0.099   & 0.095    & 0.103   & 0.083    & 0.094   & 0.094 & 0.089  & 0.102 & 0.088 & 0.108 & 0.095 & 0.083 & 0.090 & 0.082 & 0.074  &  \red{0.037}  &  0.038 & \red{0.036} \\
~      & $S_\alpha\uparrow$     & 0.716 & 0.754  & 0.701   & 0.721    & 0.705   & 0.760    & 0.727   & 0.722 & 0.725  & 0.694 & 0.742 & 0.695 & 0.716 & 0.740 & 0.723 & 0.761 & 0.787  &  \red{0.879}  &  0.868 & \red{0.878} \\
~      & $E_\phi^m\uparrow$     & 0.750 & 0.801  & 0.766   & 0.783    & 0.750   & 0.801    & 0.791   & 0.786 & 0.791  & 0.772 & 0.797 & 0.741 & 0.784 & 0.818 & 0.798 & 0.803 & 0.820  &  \red{0.911}  &  0.918 & \red{0.919} \\
~      & $\HCE_\gamma\downarrow$ & 233   & 220    & 230     & 244      & 271     & 224      & 274     & 253   & 267    & 234   & 262   & 288   & 234   & 274   & 249   & 205   & 149    & \red{103}  &   204 & \red{165} \\
\midrule
\multirow{6}*{DIS-TE2} & $maxF_\beta\uparrow$   & 0.703 & 0.755  & 0.702   & 0.712    & 0.673   & 0.756    & 0.700   & 0.720 & 0.724  & 0.681 & 0.747 & 0.680 & 0.716 & 0.743 & 0.720 & 0.759 & 0.799 &  \red{0.925}  &  0.876 & \red{0.880} \\
~      & $F^w_\beta\uparrow$    & 0.597 & 0.668  & 0.598   & 0.620    & 0.570   & 0.668    & 0.618   & 0.633 & 0.636  & 0.587 & 0.664 & 0.564 & 0.627 & 0.672 & 0.636 & 0.667 & 0.728  &  \red{0.874}  &  0.852 & \red{0.859} \\
~      & $~M~\downarrow$        & 0.107 & 0.084  & 0.102   & 0.097    & 0.109   & 0.085    & 0.099   & 0.096 & 0.092  & 0.105 & 0.087 & 0.111 & 0.095 & 0.083 & 0.092 & 0.085 & 0.070  &  \red{0.03}  &  0.035 & \red{0.034} \\
~      & $S_\alpha\uparrow$     & 0.755 & 0.786  & 0.737   & 0.755    & 0.735   & 0.788    & 0.753   & 0.761 & 0.763  & 0.729 & 0.784 & 0.740 & 0.759 & 0.777 & 0.759 & 0.794 & 0.823  &  \red{0.915}  &  0.884 & \red{0.892} \\
~      & $E_\phi^m\uparrow$     & 0.796 & 0.836  & 0.804   & 0.820    & 0.786   & 0.833    & 0.823   & 0.829 & 0.828  & 0.813 & 0.840 & 0.781 & 0.826 & 0.856 & 0.834 & 0.832 & 0.858  &  \red{0.944}  &  0.938 & \red{0.938} \\
~      & $\HCE_\gamma\downarrow$ & 474   & 480    & 501     & 542      & 574     & 490      & 593     & 567   & 586    & 516   & 555   & 621   & 512   & 600   & 556   & 451   & 340    & \red{246}  &   480 & \red{410} \\
\midrule
\multirow{6}*{DIS-TE3} & $maxF_\beta\uparrow$   & 0.748 & 0.785  & 0.726   & 0.743    & 0.699   & 0.798    & 0.730   & 0.751 & 0.747  & 0.717 & 0.784 & 0.710 & 0.752 & 0.772 & 0.745 & 0.792 & 0.830  &  \red{0.936}  &  0.885 & \red{0.898} \\
~      & $F^w_\beta\uparrow$    & 0.644 & 0.696  & 0.620   & 0.656    & 0.590   & 0.707    & 0.641   & 0.664 & 0.657  & 0.623 & 0.700 & 0.595 & 0.664 & 0.702 & 0.662 & 0.701 & 0.758  &  \red{0.89}  &  0.862 & \red{0.879} \\
~      & $~M~\downarrow$        & 0.098 & 0.083  & 0.103   & 0.092    & 0.109   & 0.079    & 0.096   & 0.092 & 0.092  & 0.102 & 0.080 & 0.109 & 0.091 & 0.078 & 0.090 & 0.079 & 0.064  &  \red{0.031}  &  0.035 & \red{0.032} \\
~      & $S_\alpha\uparrow$     & 0.780 & 0.798  & 0.747   & 0.773    & 0.748   & 0.809    & 0.766   & 0.777 & 0.774  & 0.749 & 0.805 & 0.757 & 0.780 & 0.794 & 0.771 & 0.811 & 0.836  &  \red{0.92}  &  0.883 & \red{0.896} \\
~      & $E_\phi^m\uparrow$     & 0.827 & 0.856  & 0.815   & 0.848    & 0.801   & 0.858    & 0.849   & 0.854 & 0.843  & 0.833 & 0.869 & 0.801 & 0.852 & 0.880 & 0.855 & 0.857 & 0.883  &  \red{0.954}  &  0.951 & \red{0.954} \\
~      & $\HCE_\gamma\downarrow$ & 883   & 948    & 972     & 1059     & 1058    & 965      & 1096    & 1082  & 1111   & 999   & 1049  & 1146  & 1001  & 1136  & 1081  & 887   & 687    & \red{512}  &   973 & \red{809} \\
\midrule
\multirow{6}*{DIS-TE4} & $maxF_\beta\uparrow$   & 0.759 & 0.780  & 0.729   & 0.721    & 0.670   & 0.795    & 0.699   & 0.731 & 0.725  & 0.715 & 0.772 & 0.710 & 0.749 & 0.736 & 0.731 & 0.782 & 0.827  &  \red{0.911}  &  0.848 & \red{0.874} \\
~      & $F^w_\beta\uparrow$    & 0.659 & 0.693  & 0.625   & 0.633    & 0.559   & 0.705    & 0.616   & 0.647 & 0.630  & 0.621 & 0.687 & 0.598 & 0.663 & 0.664 & 0.652 & 0.693 & 0.753  &  \red{0.857}  &  0.829 & \red{0.858} \\
~      & $~M~\downarrow$        & 0.102 & 0.091  & 0.109   & 0.107    & 0.127   & 0.087    & 0.113   & 0.107 & 0.107  & 0.111 & 0.092 & 0.114 & 0.099 & 0.098 & 0.102 & 0.091 & 0.072  &  \red{0.041}  &  0.049 & \red{0.041} \\
~      & $S_\alpha\uparrow$     & 0.784 & 0.794  & 0.743   & 0.752    & 0.723   & 0.807    & 0.744   & 0.763 & 0.758  & 0.744 & 0.792 & 0.755 & 0.776 & 0.770 & 0.762 & 0.802 & 0.830  &  \red{0.903}  &  0.854 & \red{0.874} \\
~      & $E_\phi^m\uparrow$     & 0.821 & 0.848  & 0.803   & 0.825    & 0.767   & 0.847    & 0.824   & 0.838 & 0.815  & 0.820 & 0.854 & 0.788 & 0.837 & 0.848 & 0.841 & 0.842 & 0.870  &  \red{0.944}  &  0.938 & \red{0.947} \\
~      & $\HCE_\gamma\downarrow$ & 3218  & 3601   & 3654    & 3760     & 3678    & 3653     & 3683    & 3803  & 3806   & 3709  & 3864  & 3999  & 3690  & 3817  & 3819  & 3331  & 2888   & \red{2301}  &   3799 & \red{3321} \\
\midrule
\multirow{6}*{\tabincell{c}{Overall\\DIS-TE (1-4)}} & $maxF_\beta\uparrow$   & 0.708 & 0.752  & 0.694   & 0.704    & 0.660   & 0.761    & 0.693   & 0.712 & 0.710  & 0.678 & 0.743 & 0.674 & 0.711 & 0.729 & 0.710 & 0.757 & 0.799 &  \red{0.916}  &  0.863 & \red{0.875} \\
~      & $F^w_\beta\uparrow$    & 0.603 & 0.663  & 0.590   & 0.614    & 0.554   & 0.670    & 0.608   & 0.624 & 0.620  & 0.584 & 0.658 & 0.558 & 0.622 & 0.658 & 0.628 & 0.665 & 0.726  &  \red{0.855}  &  0.839 & \red{0.856} \\
~      & $~M~\downarrow$        & 0.103 & 0.086  & 0.103   & 0.098    & 0.112   & 0.083    & 0.101   & 0.097 & 0.095  & 0.105 & 0.087 & 0.110 & 0.095 & 0.085 & 0.094 & 0.084 & 0.070  &  \red{0.035}  &  0.039 & \red{0.036} \\
~      & $S_\alpha\uparrow$     & 0.759 & 0.783  & 0.732   & 0.750    & 0.728   & 0.791    & 0.747   & 0.756 & 0.755  & 0.729 & 0.781 & 0.737 & 0.758 & 0.770 & 0.754 & 0.792 & 0.819  &  \red{0.905}  &  0.872 & \red{0.885} \\
~      & $E_\phi^m\uparrow$     & 0.798 & 0.835  & 0.797   & 0.819    & 0.776   & 0.835    & 0.822   & 0.827 & 0.819  & 0.810 & 0.840 & 0.778 & 0.825 & 0.850 & 0.832 & 0.834 & 0.858  &  \red{0.938}  &  0.936 & \red{0.939} \\
~      & $\HCE_\gamma\downarrow$ & 1202  & 1313   & 1339    & 1401     & 1395    & 1333     & 1411    & 1427  & 1442   & 1365  & 1432  & 1513  & 1359  & 1457  & 1426  & 1218  & 1016   & \red{790}  &   1364 & \red{1176} \\
        \bottomrule
    \end{tabular}
    }
    \label{tab:dis}
\end{table}

\noindent\textbf{More Qualitative Evaluation.}
\red{As shown in \cref{fig:dis_vis_page}, our method yields refined segmentation results, providing cleaner foreground masks. It also can produce precise outputs for intricate lines.}

\noindent\textbf{Cross Dataset Evaluation.}
To test the generalization ability of our model, we randomly select some images from other datasets\citep{agustsson2017ntire, lin2014microsoft, shao2019objects365} and in-the-wild images for experiments. As shown in ~\cref{fig:dis_cross}, compared to IS-Net and IS-Net-General-Use\citep{qin2022highly}, our approach exhibits finer segmentation quality across diverse images, providing cleaner foreground masks. \red{Compared to MVANet \citep{yu2024multi}, GenPercept exhibits enhanced robustness when applied to in-the-wild images. This improvement can be attributed to its large-scale pre-training on the LAION dataset and the extensive parameterization of the diffusion model. In contrast, the backbone of MVANet is pre-trained on the ImageNet dataset \citep{deng2009imagenet}, which may limit its performance in more diverse environments.}

It is noteworthy that IS-Net-General-Use is fine-tuned on extra datasets to enhance generalization, which indicates that our method has a stronger generalization ability.

\subsection{Image Matting}

\noindent\textbf{Implementation Details.}
We utilize P3M10K \citep{li2021privacy}, the largest portrait matting dataset with high-resolution portrait images along with high-quality alpha masks to train our model. The training set contains 9,421 high-quality images and annotations and the test set P3M-500-NP is composed of 500 public images from the Internet. We train our GenPercept with pixel-wise MSE loss to further improve the final performance.

\noindent\textbf{More Qualitative Evaluation. } In \cref{fig:matting_supplementary}, we showcase more qualitative results for the image matting task. It is worth noting that our model works well in various resolutions, light environments, human poses, and human orientations. \red{More importantly, our GenPercept model trained on human matting images shows much more robustness to other objects compared to existing P3M10K \citep{li2021privacy} SOTA method ViTAE-S \citep{ma2023rethinking}, as illustrated in \cref{fig:image_matting_robust_supp}. It shows that the ViTAE-S overfits the human matting task, while GenPercept preserves the generalization ability. Besides, we also train GenPercept on a more general image matting task on the Composition-1K \citep{xu2017deep} dataset. As shown in \cref{fig:image_matting_transparent_supp}, GenPercept shows robustness on more types of objects such as semi-transparent objects, hollow objects, etc.}

\subsection{Image Segmentation}
\noindent\textbf{More Qualitative Evaluation. } In \cref{fig:supp_rgb_seg}, we showcase more qualitative results for the image segmentation task. Our method shows much robustness on the trained categories of complex in-the-wild images. Due to the limited annotation categories and little negative label of “unknown category”, it sometimes fails in outdoor scenes and unseen categories such as cats and cars. Please zoom in for better visualization and more details.

\subsection{Human Pose Estimation}
\noindent\textbf{More Qualitative Evaluation. } In \cref{fig:human_pose},  we showcase more qualitative results for the human pose estimation. We conduct experiments on MHP dataset \citep{li2017multiplehuman}, and we use mmpose \citep{mmpose2020} to render the human pose following the setting of \citep{bai2023sequential}.

\newpage

\begin{figure}[H]
    \centering
    \includegraphics[width=\linewidth]{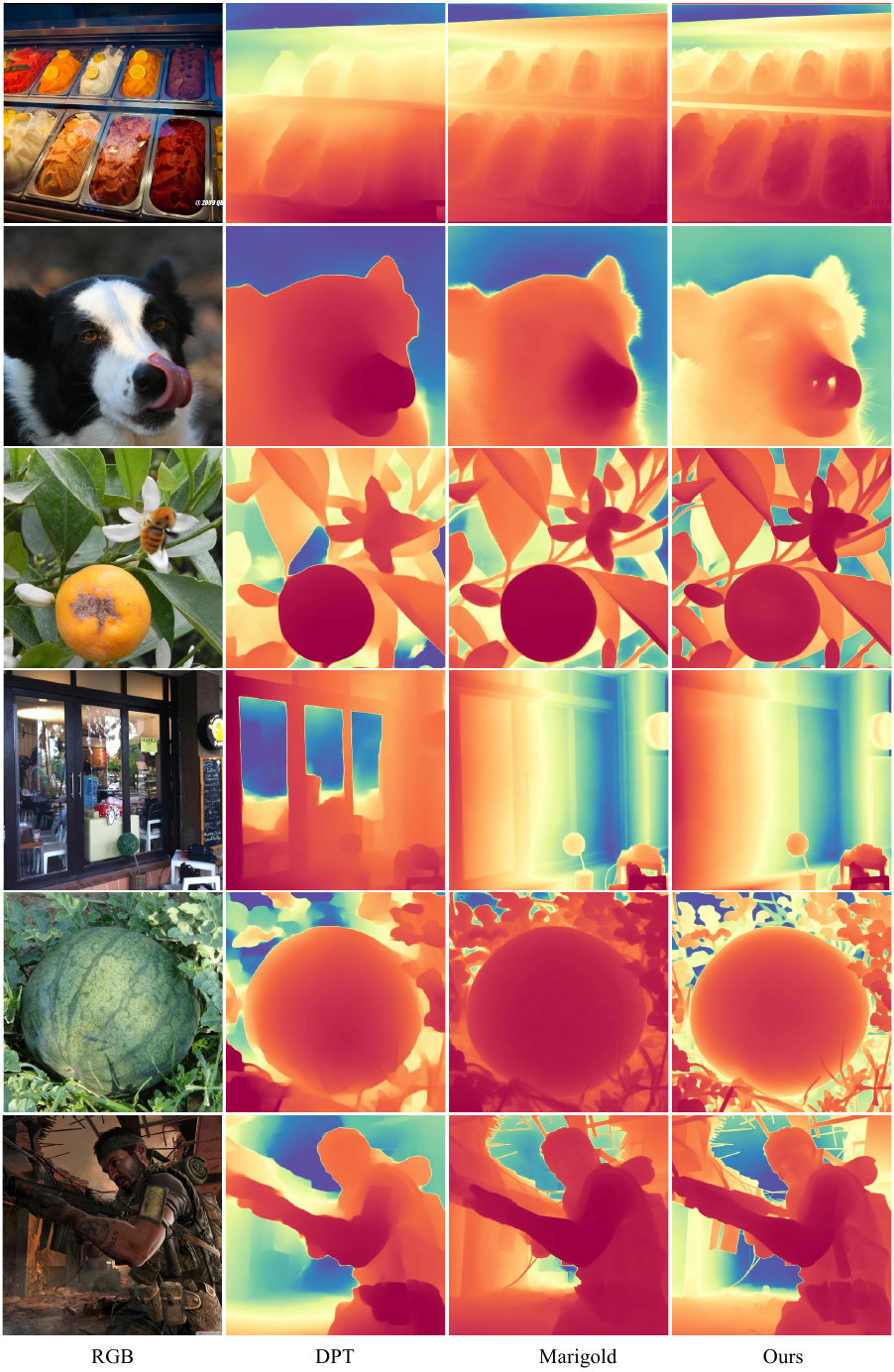}
\end{figure}

\begin{figure}[H]
    \centering
    \includegraphics[width=.9\linewidth]{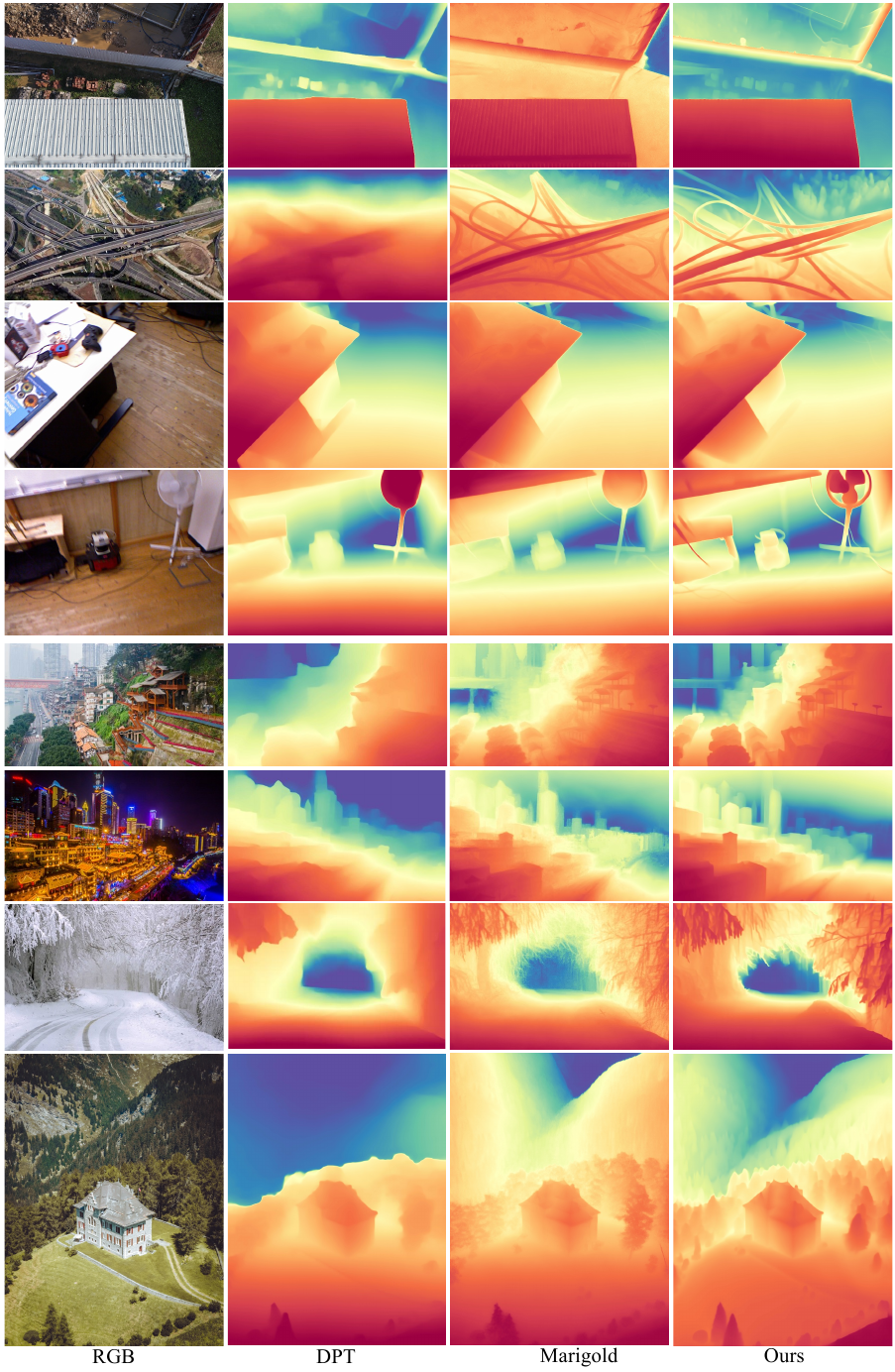}
    \caption{More qualitative results for monocular depth estimation.}
    \label{fig:supp_depth}
\end{figure}

\begin{figure}[H]
    \centering
    \includegraphics[width=\linewidth]{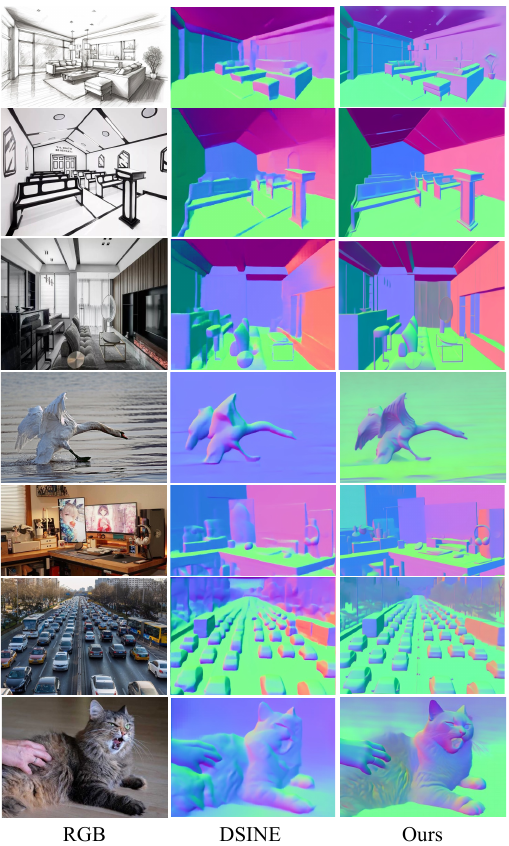}
\end{figure}

\begin{figure}[H]
    \centering
    \includegraphics[width=\linewidth]{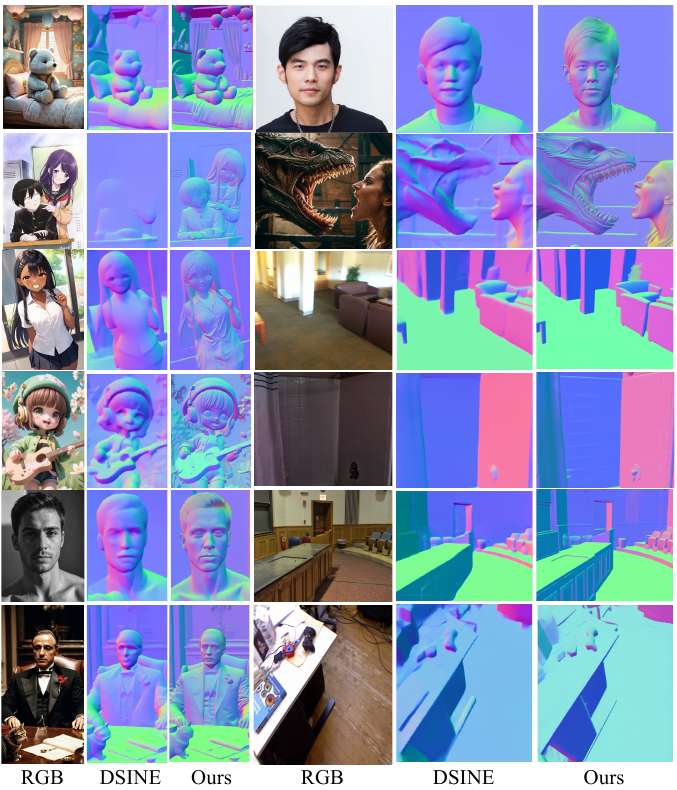}
    \caption{More qualitative comparisons of surface normal estimation. Our GenPercept can achieve more detailed results, even compared to the competitive CVPR2024 method DSINE \citep{bae2024rethinking}. Note that these two visualization coordinate systems are slightly different.}
    \label{fig:supp_normal}
\end{figure}

\begin{figure}[H]
    \centering
    \includegraphics[width=\textwidth]{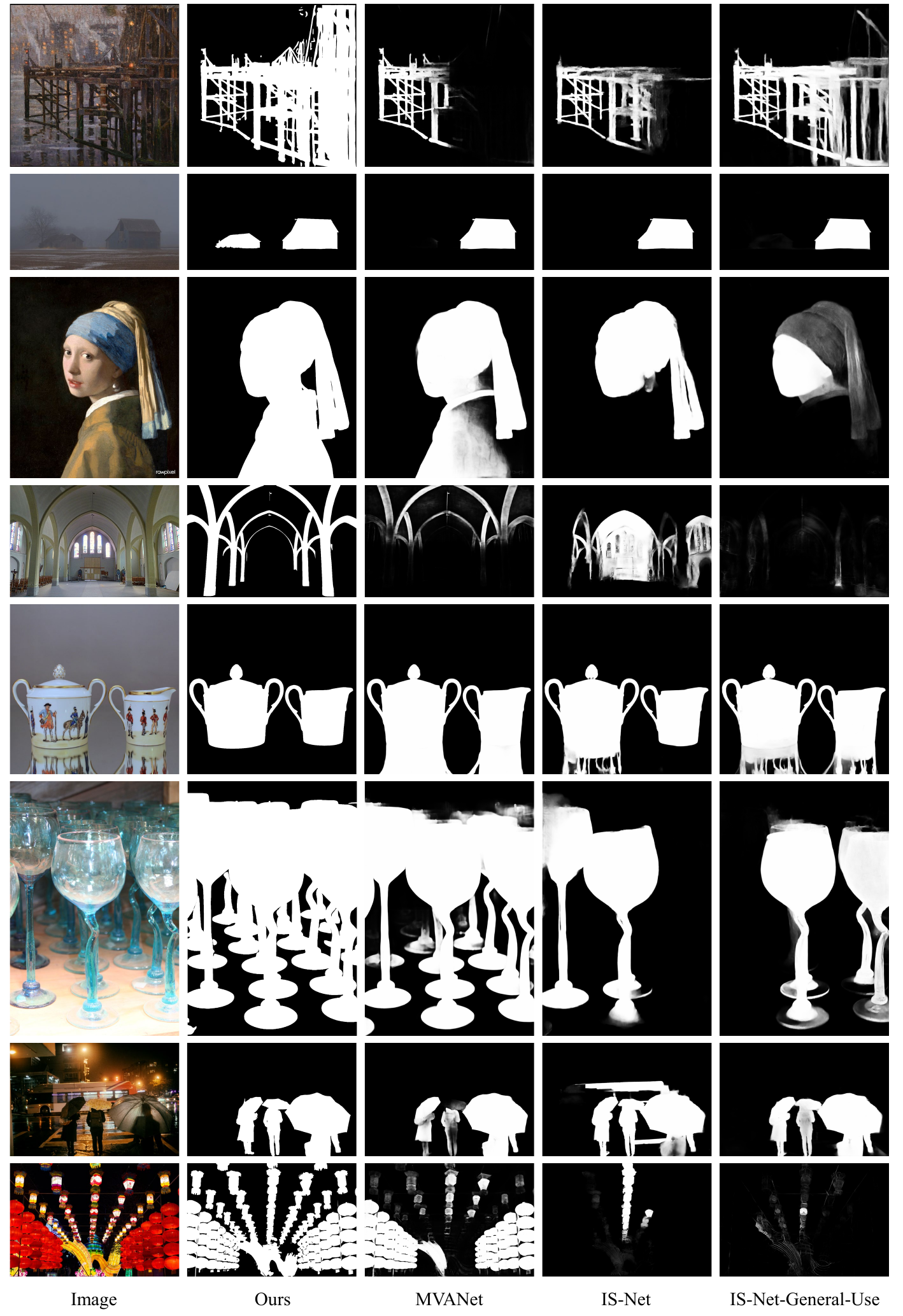}
\end{figure}

\begin{figure}[H]
    \centering
    \includegraphics[width=\textwidth]{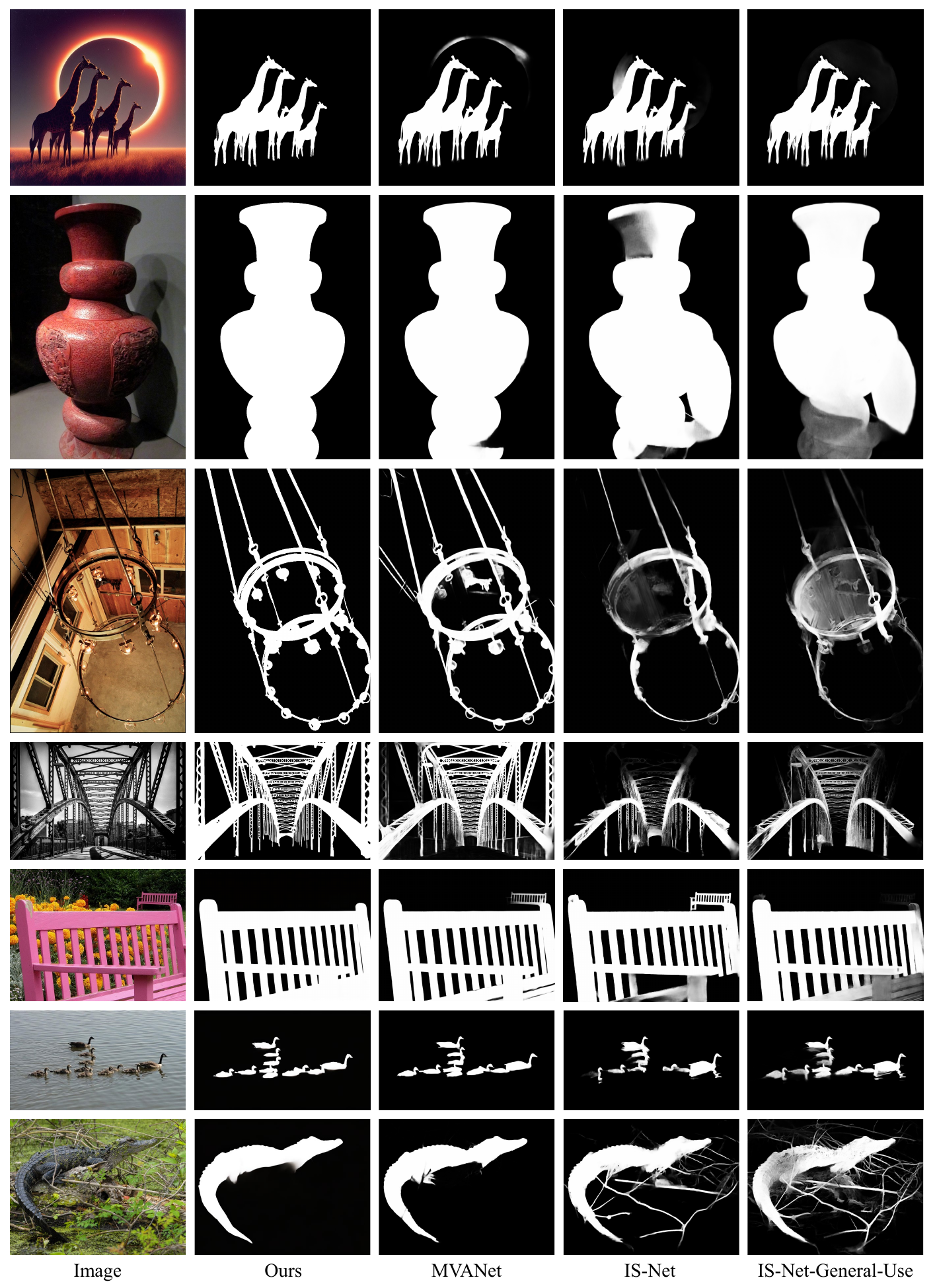}
    \vspace{-1 em}
    \caption{\red{Cross dataset comparison of our model and other models.}}
    \label{fig:dis_cross}
\end{figure}

\begin{figure}[H]
    \centering
    \includegraphics[width=\linewidth]{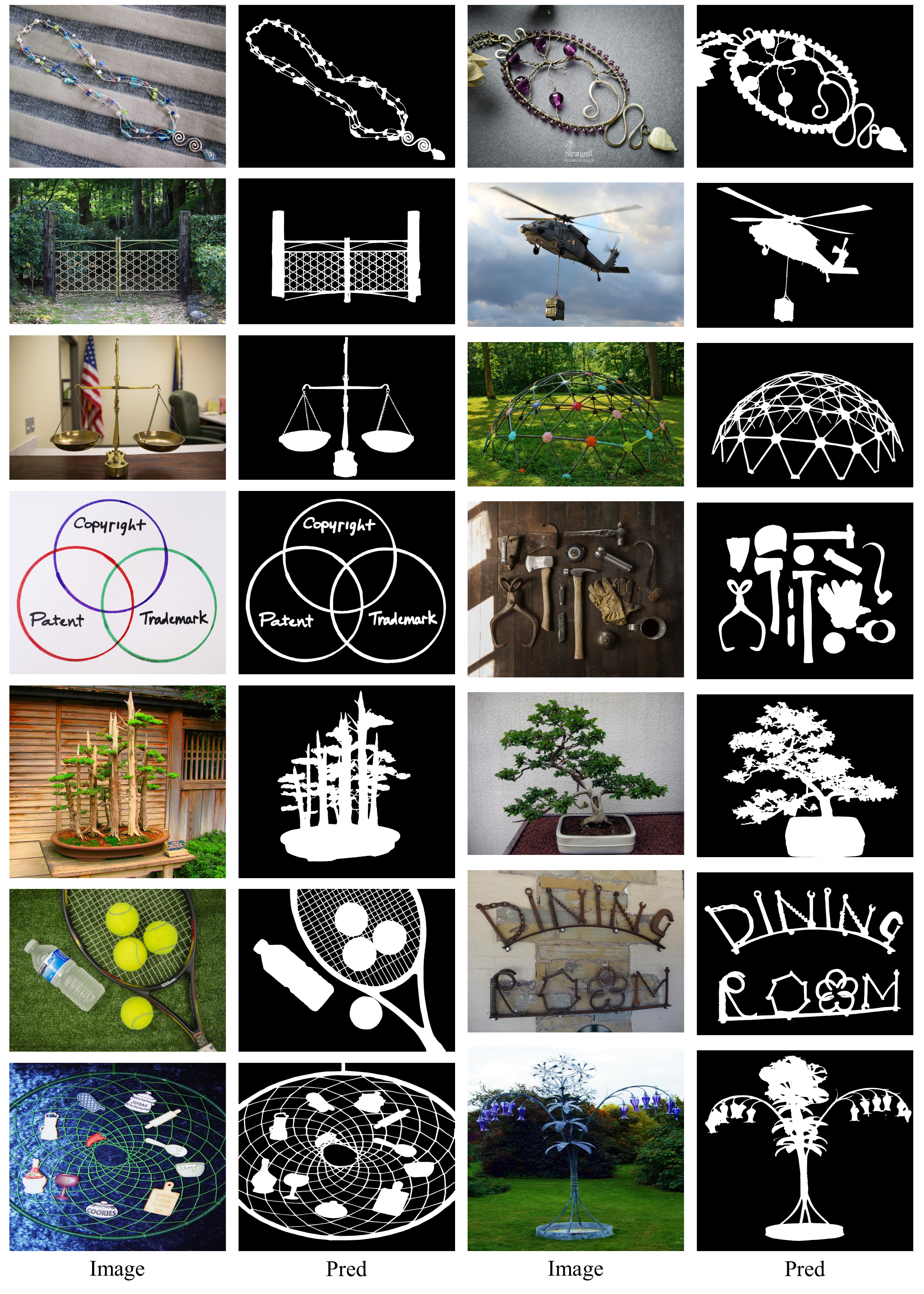}
    \caption{Qualitative results of dichotomous image segmentation.}
    \label{fig:dis_vis_page}
\end{figure}

\begin{figure}[H]
    \centering
    \includegraphics[width=\linewidth]{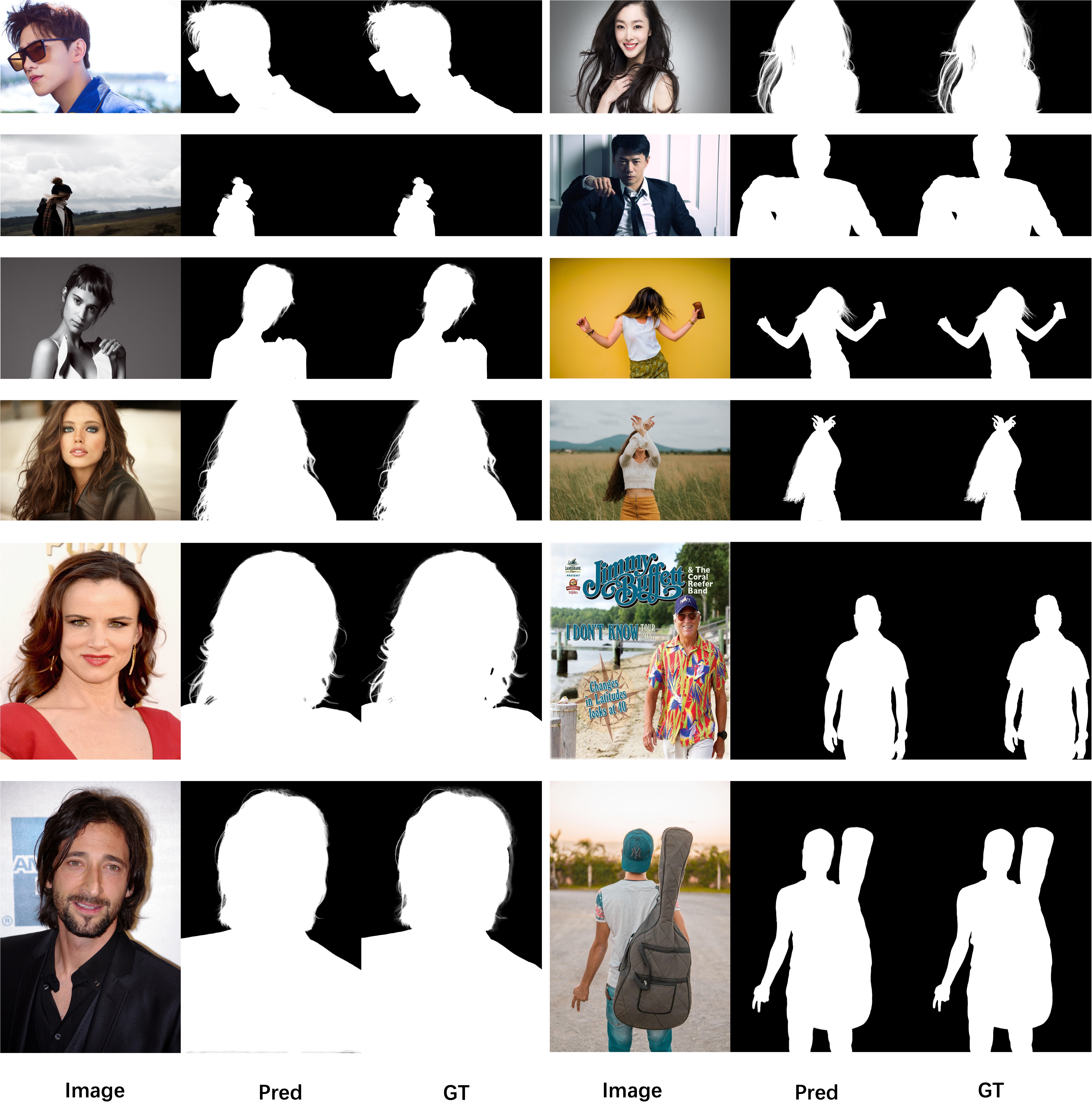}
    \caption{More qualitative results for image matting.}
    \label{fig:matting_supplementary}
\end{figure}

\begin{figure}[H]
    \centering
    \includegraphics[width=\linewidth]{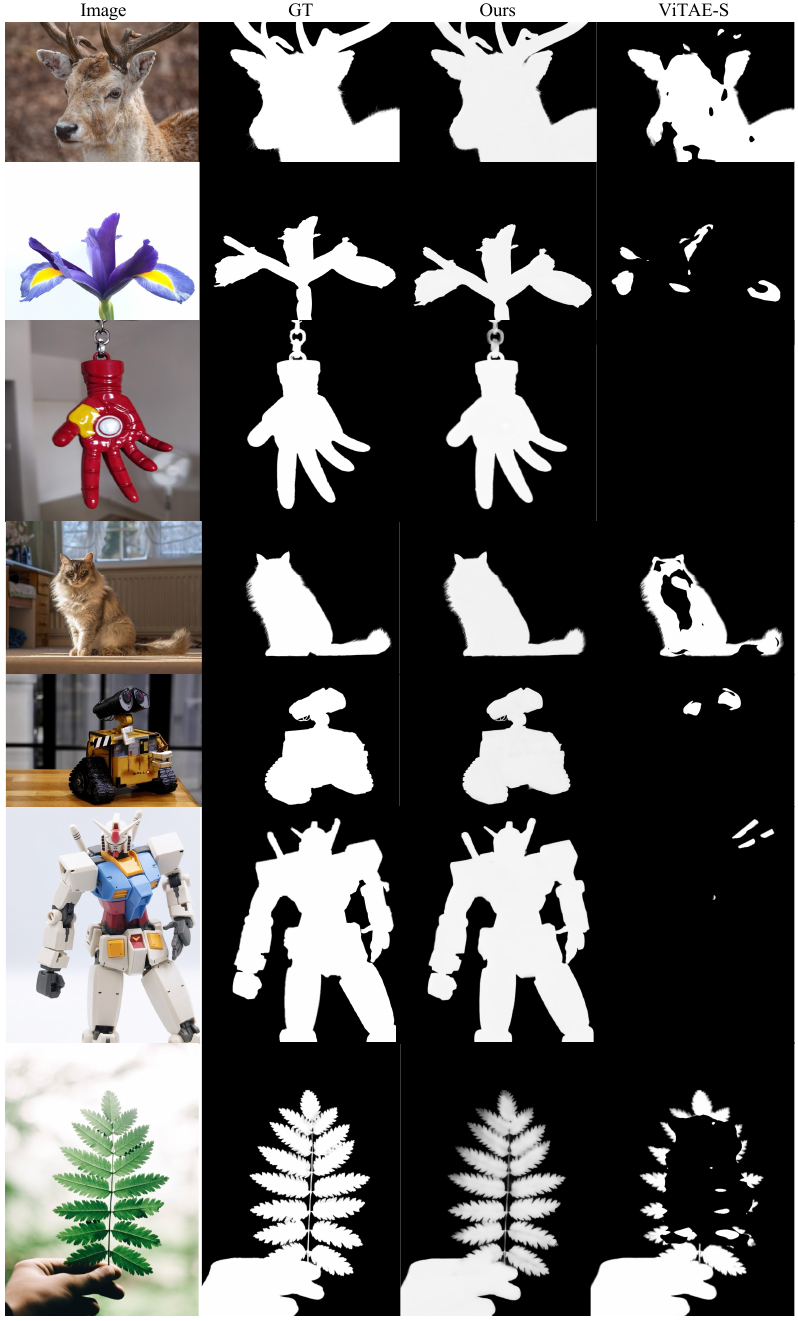}
\end{figure}

\begin{figure}[H]
    \centering
    \includegraphics[width=\linewidth]{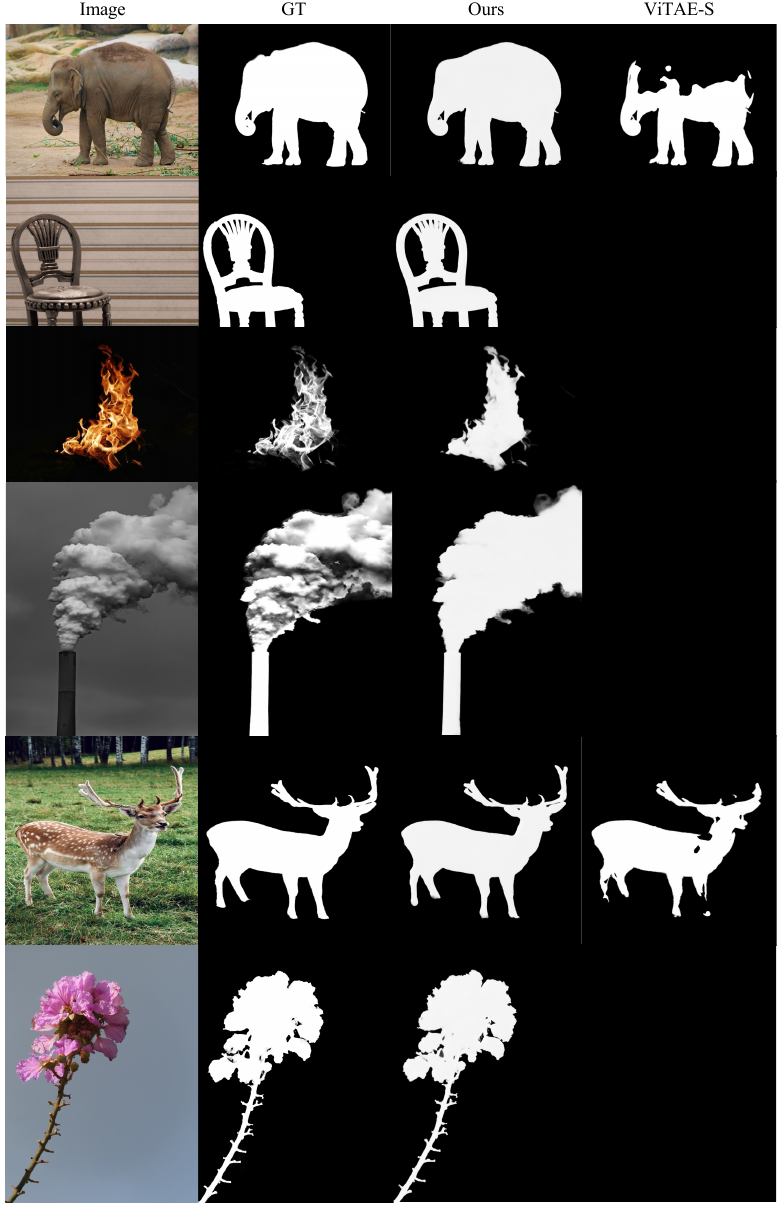}
    \caption{\red{Generalization ability of the human matting model to general image matting images.}}
    \label{fig:image_matting_robust_supp}
\end{figure}

\begin{figure}[H]
    \centering
    \includegraphics[width=.85\linewidth]{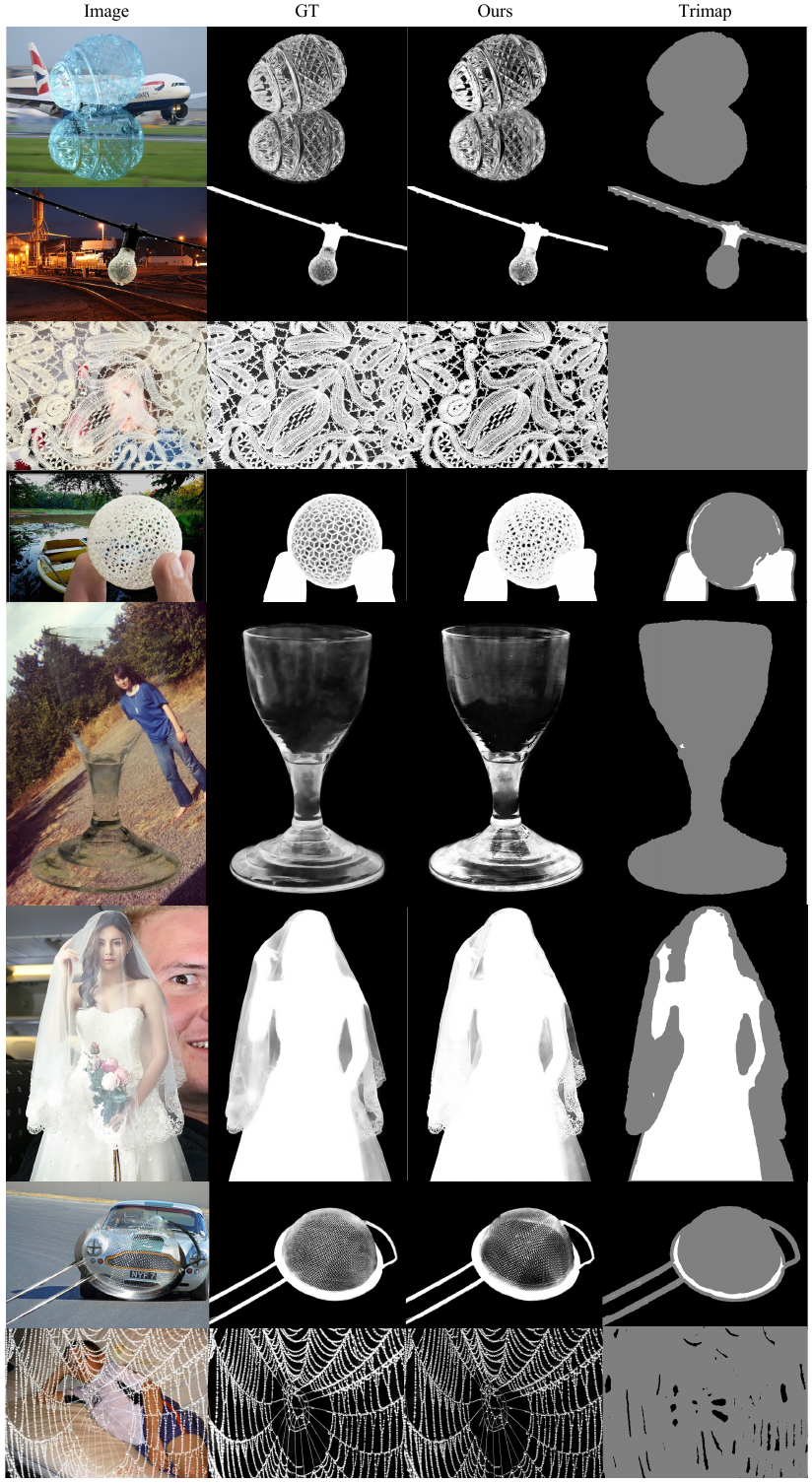}
    \caption{\red{More types of image matting such as semi-transparent objects.}}
    \label{fig:image_matting_transparent_supp}
\end{figure}

\begin{figure}[H]
    \centering
    \includegraphics[width=\linewidth]{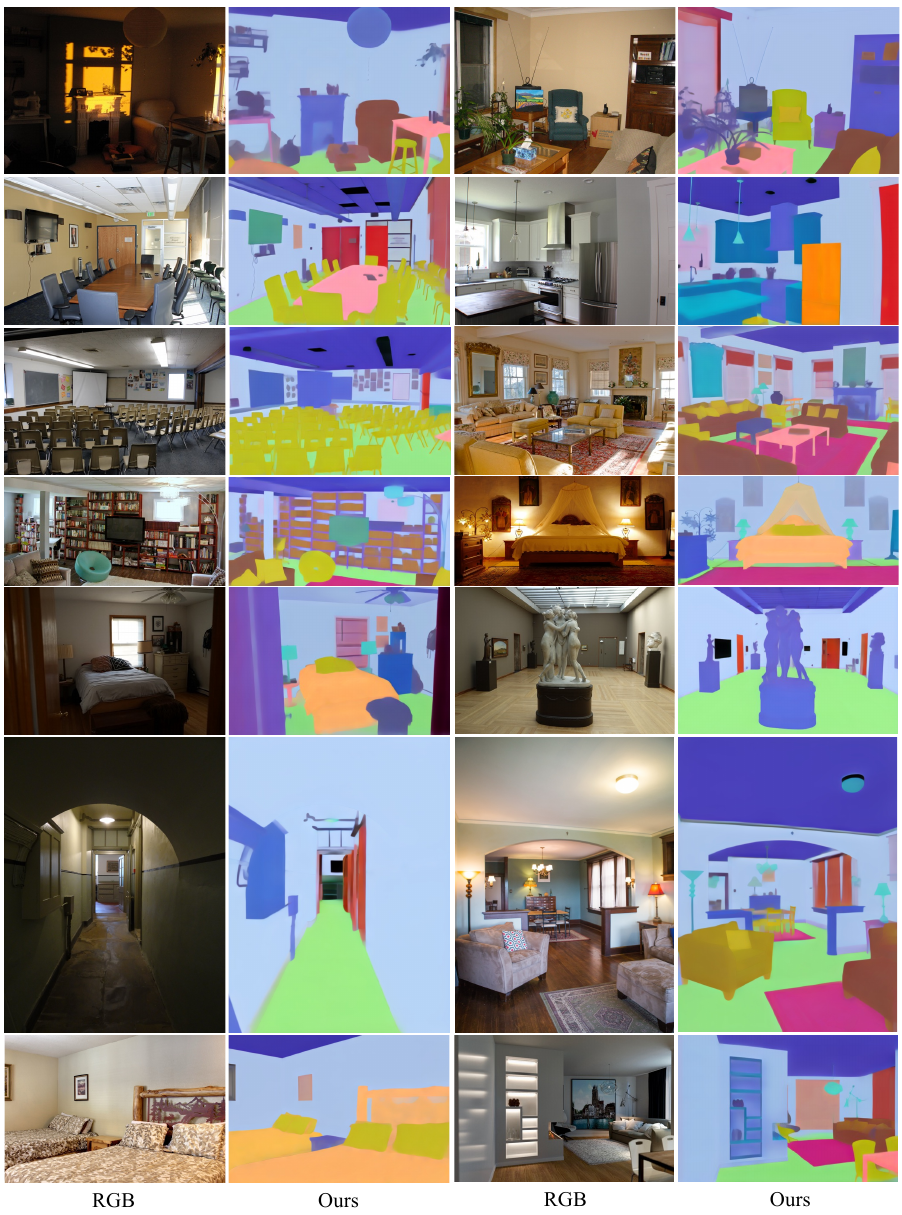}
    \caption{More qualitative results for image segmentation.}
    \label{fig:supp_rgb_seg}
\end{figure}

\begin{figure}[H]
    \centering
    \includegraphics[width=\linewidth]{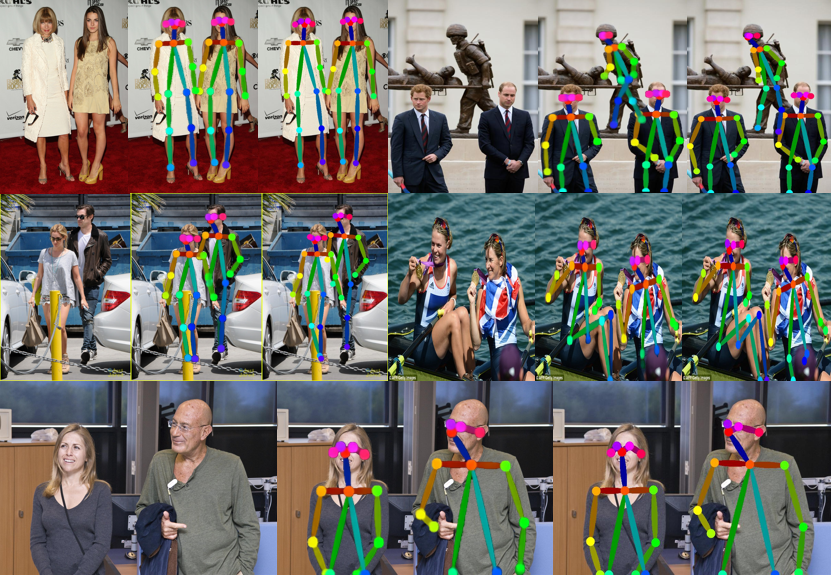}
    \caption{More qualitative results for human pose estimation. (Left: original Image, Mid: prediction, Right: ground truth)}
    \label{fig:human_pose}
\end{figure}

\bibliography{cvml,iclr2025_conference}
\bibliographystyle{iclr2025_conference}

\end{document}